\PassOptionsToPackage{table}{xcolor}
\documentclass[letterpaper]{article} 
\usepackage[preprint]{aaai2027}  
\usepackage[hyphens]{url}  
\usepackage{graphicx} 
\usepackage{natbib}  
\usepackage{caption} 
\usepackage{algorithm}
\usepackage{algorithmic}

\usepackage{newfloat}
\usepackage{listings}
\DeclareCaptionStyle{ruled}{labelfont=normalfont,labelsep=colon,strut=off} 
\floatstyle{ruled}
\newfloat{listing}{tb}{lst}{}
\floatname{listing}{Listing}

\usepackage{booktabs}

\usepackage[utf8]{inputenc}
\usepackage[T1]{fontenc}
\usepackage{algorithm}
\usepackage{algorithmic}
\usepackage{graphicx}
\usepackage{booktabs}
\usepackage{multirow}
\usepackage{amsmath}
\usepackage{amsfonts}
\usepackage{enumitem}
\usepackage{url} 
\usepackage{xspace}
\usepackage{xcolor}
\usepackage[most]{tcolorbox}
\usepackage{epigraph}
\usepackage{amsthm}
\usepackage{amssymb}
\usepackage{arydshln}

\usepackage{pifont}
\newcommand{\fourstarblack}{\ding{70}} 
\newcommand{\fourstarwhite}{\ding{71}} 

\definecolor{langlightblue}{rgb}{0.3, 0.65, 1}
\definecolor{langblue}{rgb}{0, 0.4, 0.8}
\definecolor{langmildblue}{rgb}{0.0, 0.45, 0.73}
\definecolor{langdarkblue}{rgb}{0.0, 0.0, 0.61}
\definecolor{langred}{rgb}{0.81, 0.09, 0.13}
\definecolor{langlightgreen}{rgb}{0.80, 0.97, 0.85}
\definecolor{langgreen}{rgb}{0.18, 0.55, 0.34}
\definecolor{langdarkgreen}{rgb}{0.0, 0.45, 0.38}
\definecolor{bingpink}{rgb}{1.0, 0.41, 0.71}
\definecolor{MidnightBlue}{RGB}{37,101,147}
\definecolor{gptcommentcolor}{RGB}{180,32,37}
\definecolor{auditcommentcolor}{RGB}{35,88,168}
\newcommand{\gain}[1]{\textsubscript{\textbf{\textit{\textcolor{langdarkgreen}{#1}}}}}

\usepackage{titletoc}

\newtcblisting{promptbox}[2][]{
    enhanced,
    width=\linewidth,
    colback=gray!5,
    colframe=gray,
    title=\textbf{#2},
    listing only,
    listing options={
        basicstyle=\normalfont\itshape\small,
        breaklines=true,
        breakatwhitespace=true,
        columns=fullflexible
    },
    #1
}

\title{GuideSkill: Evolving Executable LLM Agent Skills for \\ Guideline-Grounded Clinical Reasoning}

\author{
    Lang Cao\quad
    Yuhao Shen\textsuperscript{\fourstarblack}\quad
    Tianyang Luo\quad
    Simo Du\textsuperscript{\fourstarwhite}\quad
    Hao Peng\quad
    Yue Guo 
}
\affiliations{
    University of Illinois Urbana-Champaign\\
    \textsuperscript{\fourstarblack}The Chinese University of Hong Kong, Shenzhen\\
    \textsuperscript{\fourstarwhite}Albert Einstein College of Medicine
}

\begin{document}

\maketitle
\pagestyle{plain}
\thispagestyle{plain}

\begin{abstract}
Clinical practice guidelines (CPGs) encode diagnostic criteria, but LLM systems typically retrieve guideline text or absorb it through training rather than execute its rules. We introduce \textit{GuideSkill}, an external reasoning layer that compiles disease-specific criteria into executable functions returning ordinal diagnostic-support scores. \textit{GuideSkill-Zero} is initialized from guidelines, while \textit{GuideSkill-Evo} uses case--diagnosis pairs to refine covered skills and add missing diagnoses. At inference, an LLM proposes a differential diagnosis, grounds the features required by each matched skill, and fuses its ranking with the executed skill scores. Across four benchmarks and four backbones, \textit{GuideSkill-Zero} improves macro-average accuracy over guideline RAG by 13.45\% on average. \textit{GuideSkill-Evo} achieves the highest macro-average for every backbone, improves over direct inference by 18.49\% relatively, and increases gold-label skill coverage from 56.5\% to 99.5\%. On Qwen3.5-9B, it also exceeds the strongest parameter-update baseline by 11.16\% without updating the backbone. Expert evaluation further indicates that \textit{GuideSkill} produces clinically sound and broadly acceptable skills, suggesting that its initialized and evolved rules are reliable and practically meaningful. These results support executable skills as a model-agnostic mechanism for combining guideline-derived procedures with case-derived diagnostic patterns.
\end{abstract}


\section{Introduction}

Clinical diagnosis is not only a knowledge-recall problem. It is a multistep process that requires gathering and synthesizing patient information, generating and comparing plausible diagnoses, determining which findings and thresholds support or weaken each candidate, excluding alternatives, and identifying decisive tests~\cite{mcduff2025differential,hager2024evaluation,cao2026ehr,you2026improving}. Clinical practice guidelines (CPGs) encode evidence-based recommendations and conditional decision logic that can guide these decisions~\cite{iom2011guidelines,shen2026medguidex}. However, providing guideline-derived criteria to an LLM does not guarantee that they will be applied correctly to the patient: even when guideline-based checklists are supplied, criterion-level evaluation remains imperfect~\cite{schubert2025guideline}. LLMs can generate ranked differential diagnoses from free-text cases~\cite{mcduff2025differential}, but may omit clinician-identified reasoning evidence and fail to follow diagnostic guidelines~\cite{wu2025medcasereasoning,hager2024evaluation}. We therefore ask whether CPGs can be transformed from passive references into an evolvable library of executable diagnostic skills that combines flexible candidate generation with explicit disease-specific rule application.

Prior work incorporates CPGs by supplying guideline text or checklists at inference~\cite{schubert2025guideline}, adapting model parameters with guideline-containing corpora or guideline-derived supervision~\cite{chen2023meditron,staniek2025training,shen2026medguidex}, and translating guidelines into structured decision trees or program-aided pathways~\cite{oniani2024enhancing,li2023meddm,deng2026cpgprompt}. These approaches demonstrate several ways to operationalize guideline knowledge in LLM systems, but they leave open a complementary systems question: can guideline-derived procedures be organized as an external, disease-indexed skill library that compares evidence across candidate diagnoses, transfers across LLM backbones, and expands without updating model parameters? This question is particularly relevant to differential diagnosis, where several conditions may plausibly explain the same presentation and must be compared against disease-specific evidence~\cite{mcduff2025differential}. Because the initial guideline corpus covers only a finite set of diagnoses, we further examine whether labeled cases can extend the library to diagnoses absent from that corpus, including the rare and complex conditions represented in case-report benchmarks~\cite{wu2025medcasereasoning}.

We introduce \textit{GuideSkill}, which transforms CPGs from passive references into an external library of executable, disease-specific skills. Externalizing this knowledge makes diagnostic criteria reusable across LLM backbones and allows the library to be updated without modifying model parameters. \textit{GuideSkill-Zero} initializes the library from guidelines, providing explicit supporting and contradictory rules for covered diagnoses. \textit{GuideSkill-Evo} then uses labeled cases to refine existing skills and add diagnoses absent from the guideline corpus, expanding coverage while updating disease-specific decision rules outside the backbone.
At inference, the LLM and the skill library address complementary limitations. The LLM generates a ranked differential from the full patient narrative, including diagnoses beyond the finite skill library, while matched skills apply explicit disease-specific criteria to covered candidates. Their fusion preserves open-ended candidate generation while allowing reusable evidence rules to refine the ranking, rather than relying entirely on either an LLM-only ranking or incomplete skill coverage. Because the skills remain external, the same diagnostic logic can be inspected, updated, and reused across backbones.

We evaluate \textit{GuideSkill} on four heterogeneous diagnostic-reasoning benchmarks—MedCaseReasoning~\cite{wu2025medcasereasoning}, ER-Reason~\cite{mehandru2025er}, MIMIC-CDM-FI~\cite{hager2024evaluation}, and MedThink-Bench~\cite{zhou2025automating}—using four proprietary and open-weight LLM backbones. Using only guideline-derived executable skills, \textit{GuideSkill-Zero} achieves higher macro-average accuracy than guideline RAG for every backbone. After evolution, \textit{GuideSkill-Evo} outperforms direct inference in all 16 dataset--backbone comparisons, with a mean relative improvement of 18.49\%, while increasing pooled gold-label skill coverage from 56.50\% to 99.50\%. On MedThink-Bench, which is excluded from evolution training, it achieves the best or tied-best accuracy across all four backbones, demonstrating transfer to an unseen benchmark. On Qwen3.5-9B, it also outperforms the evaluated supervised fine-tuning, reinforcement-learning, and guideline-decision-tree baselines without updating the backbone. Together, these results demonstrate the benefits of external executable skills for diagnostic accuracy and coverage.

Our contributions are threefold:
\begin{itemize}[leftmargin=*, itemsep=0pt, labelsep=5pt, topsep=0pt]
    \item We formulate guideline-grounded diagnosis as agentic skill execution and develop a disease-indexed compilation pipeline that converts CPG recommendations into executable ordinal scorers.
    \item We introduce a case-conditioned evolution mechanism that refines covered skills and adds previously uncovered diagnoses outside the LLM parameters.
    \item We develop a candidate-level inference procedure and evaluate it across four backbones and four benchmarks, separating the contribution of guideline-only initialization from the additional coverage and downstream gains obtained through evolution.
\end{itemize}

\section{Related Work}

\noindent \textbf{Clinical Guidelines as Reasoning Resources.}
Because CPGs synthesize reviewed clinical evidence into recommendations, they are a valuable foundation for clinical decision support~\cite{iom2011guidelines}. LLM-based systems have incorporated them as retrieved prompt context~\cite{schubert2025guideline,oniani2024enhancing}, supervised examples~\cite{staniek2025training}, or reinforcement-learning signals~\cite{tziakouri2025reinforcement}. More structured approaches operationalize their decision logic: MedDM represents clinical pathways as LLM-executable guidance trees~\cite{li2023meddm}, CPGPrompt translates CPGs into decision trees traversed during inference~\cite{deng2026cpgprompt}, and MedGuideX executes guideline logic to generate factual and counterfactual supervision for post-training~\cite{shen2026medguidex}. However, the guideline-derived component of these systems remains bounded by the source corpus: it provides no procedure for diagnoses outside that corpus and no mechanism to learn additional diagnostic patterns from labeled case presentations. \textit{GuideSkill} addresses this gap by initializing an external skill library from guidelines and using labeled cases to refine covered skills and add uncovered diagnoses. It thereby retains guideline-derived procedures while extending beyond the initial corpus.
\\

\noindent \textbf{Evolving Skills in Healthcare and Biomedicine.}
Agent skills externalize reusable procedures so that capabilities can accumulate without changing model parameters. Trace2Skill induces transferable operating procedures from recurring patterns in execution trajectories~\cite{ni2026trace2skill}, while SkillClaw aggregates cross-user trajectories to refine and extend a shared skill repository~\cite{ma2026skillclaw}. In healthcare, an empirical study of public skills finds that they primarily support patient-facing workflow automation and monitoring, with limited coverage of diagnosis and treatment~\cite{xu2026empirical}. Related biomedical systems target scientific workflows: SkillFoundry mines heterogeneous resources into validated executable skills and demonstrates them on genomics tasks~\cite{shen2026skillfoundry}, whereas STELLA evolves reasoning templates and tool use for biomedical research and experimental discovery~\cite{jin2025stella}. These studies capture experience or operational procedures but do not address evolving evidence-based knowledge for patient-level diagnosis. \textit{GuideSkill} fills this gap by initializing skills from CPGs, expanding them with labeled cases, and executing them against patient evidence.

\begin{figure*}[t!]
\centering
\includegraphics[width=\textwidth]{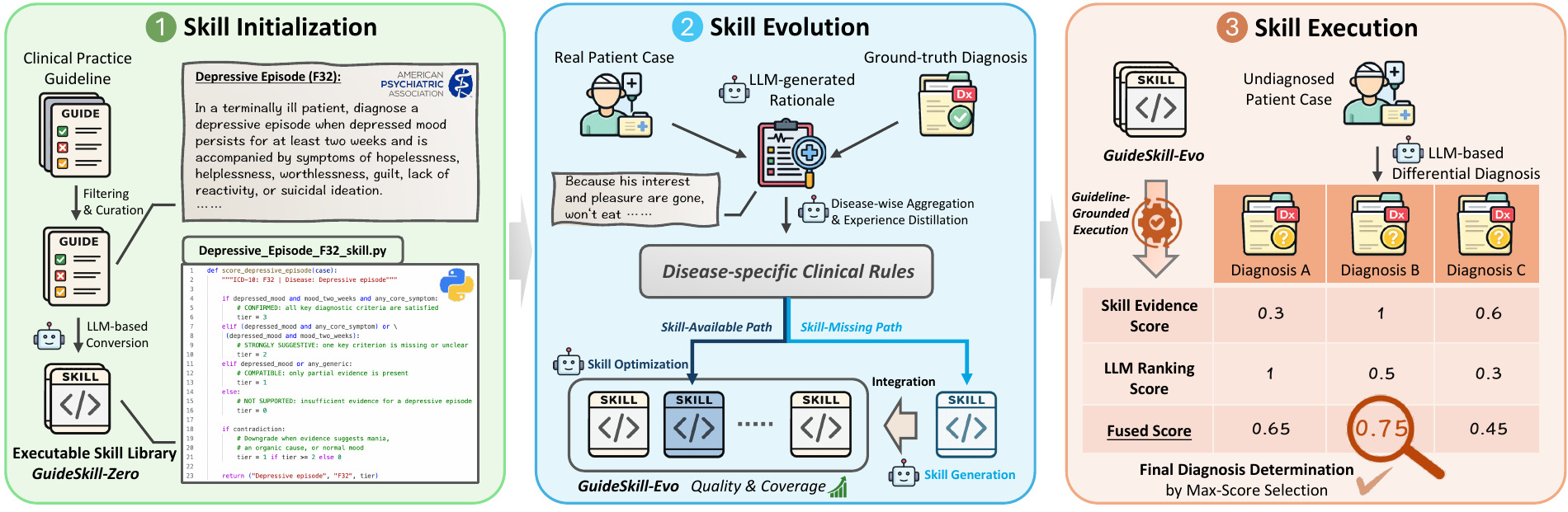}
\caption{Overview of \textit{GuideSkill}. Guideline recommendations are compiled into the initial executable library (\textit{GuideSkill-Zero}); labeled cases refine covered skills and add missing diagnoses (\textit{GuideSkill-Evo}); and, during inference, executed skill scores are fused with an LLM-generated differential ranking.}
\label{fig:framework}
\end{figure*}

\section{Methodology}

Figure~1 provides an overview of \textit{GuideSkill}. The framework first constructs an initial skill library, \textit{GuideSkill-Zero}, from clinical guidelines, then evolves the library with real patient cases to obtain \textit{GuideSkill-Evo}, and finally applies the learned disease-specific skills during inference.

\subsection{Task Formulation}

Given an undiagnosed patient case $x$, our goal is to predict the final diagnosis $d^\star \in \mathcal{D}$. Here, $x$ includes the patient’s demographics, chief complaint, medical history, physical examination findings, and available test results. The diagnosis space $\mathcal{D}$ is standardized using three-character WHO ICD-10 categories~\cite{who2019icd10}, which provide normalized disease identifiers across datasets.

During inference, an LLM $M$ first generates a ranked differential diagnosis set:
\begin{equation}
\mathcal{C}(x)=\{d_1,\dots,d_K\}, \quad \mathcal{C}(x)\subseteq \mathcal{D},
\end{equation}
where $K=5$ by default. For each candidate diagnosis $d \in \mathcal{C}(x)$, we convert its rank into an LLM ranking score $s_{\mathrm{LLM}}(x,d)$. Specifically, if $d$ is ranked at position $r_d$ with zero-based indexing, then
\begin{equation}
s_{\mathrm{LLM}}(x,d)=\frac{1}{r_d+1}.
\end{equation}

\textit{GuideSkill} then retrieves the corresponding disease-specific skill $g_d$ from the skill library $\mathcal{G}$ and executes it on the patient case to obtain a skill-based evidence score:
\begin{equation}
s_{\mathrm{skill}}(x,d)=g_d(x).
\end{equation}
The final diagnosis score combines the LLM ranking score and the skill-based evidence score:
\begin{equation}
S(x,d)=
\alpha s_{\mathrm{LLM}}(x,d)
+
(1-\alpha)s_{\mathrm{skill}}(x,d),
\end{equation}
where $\alpha \in [0,1]$ controls the relative contribution of the two signals. The final prediction is selected from the candidate set:
\begin{equation}
\hat{d}=
\arg\max_{d \in \mathcal{C}(x)}
S(x,d).
\end{equation}

\subsection{Skill Initialization}


The first stage constructs the initial skill library, denoted as \textit{GuideSkill-Zero}. Given a collection of clinical practice guidelines,
$\mathcal{P}=\{p_1,\dots,p_M\},$
we first filter and curate the raw guideline documents to obtain a high-quality guideline set $\mathcal{P}'$. Details of the guideline preprocessing are provided in Appendix B. Each curated guideline is then parsed into disease-specific clinical recommendations, where each recommendation captures a clinically relevant diagnostic criterion, finding, exclusion rule, or decision rule that supports or refutes a specific diagnosis.

Since \textit{GuideSkill} operates at the ICD-10 category level, we group recommendations by their mapped ICD-10 diagnosis category. For a diagnosis category $d$, we denote the extracted recommendation set as:
$\mathcal{R}_d =
\{r_{d,1}, r_{d,2}, \dots, r_{d,m_d}\},$
where $m_d$ is the number of recommendations extracted and mapped to diagnosis category $d$.

Each recommendation set $\mathcal{R}_d$ is then converted by an LLM into executable diagnostic logic, yielding an initial disease-specific skill:
\begin{equation}
g_d^{(0)} =
\mathrm{LLM}_{\mathrm{Compile}}(\mathcal{R}_d).
\end{equation}
The initialized skill library is therefore defined as:
\begin{equation}
\mathcal{G}^{(0)} =
\{g_d^{(0)} \mid d \in \mathcal{D}_{\mathrm{guide}}\},
\end{equation}
where $\mathcal{D}_{\mathrm{guide}}$ denotes the set of ICD-10 diagnosis categories covered by the curated guidelines.

Each skill $g_d^{(0)}$ is implemented as an executable Python function. It takes the relevant features of a patient case $x$ as input and returns a skill-based evidence score:
\begin{equation}
s_{\mathrm{skill}}(x,d) =
g_d^{(0)}(x).
\end{equation}
The score is computed by matching the clinical evidence in $x$ against the executable diagnostic rules encoded in $g_d^{(0)}$. Specifically, each skill assigns the evidence to one of four support levels: \texttt{Confirmed}, \texttt{Strongly Suggestive}, \texttt{Compatible}, and \texttt{Not Supported}. Contradictory evidence is further used to downgrade the support level when applicable. The final evidence score is normalized to the range $[0,1]$.
Since each skill is executable, it can be directly invoked by an LLM agent to obtain a structured evidence score for diagnosis $d$. In this way, \textit{GuideSkill-Zero} provides a guideline-grounded and interpretable mechanism for diagnosis scoring.

\subsection{Skill Evolution}

Although \textit{GuideSkill-Zero} is grounded in clinical guidelines, guidelines mainly capture general diagnostic principles and key decision points, while real-world cases often contain heterogeneous presentations and atypical clinical patterns. We therefore evolve the skill library using labeled patient cases. Let the training set be $\mathcal{T}=\{(x_j,d_j)\}_{j=1}^{n}$, where $x_j$ is a patient case and $d_j$ is the ground-truth diagnosis.

For each case $(x_j,d_j)$, the LLM generates a diagnostic rationale $z_j$ that summarizes the clinical evidence supporting $d_j$. Rationales associated with the same diagnosis are then aggregated and distilled into additional disease-specific rules:
\begin{equation}
\Delta \mathcal{R}_d =
\mathrm{LLM}_{\mathrm{Distill}}\big(\{z_j \mid d_j=d\}\big).
\end{equation}

The newly distilled rules are used to update the skill library. If diagnosis $d$ already has an existing skill, the rules are used to optimize that skill; otherwise, they are used to generate a new skill:
\begin{equation}
g_d^{(t+1)} =
\begin{cases}
\mathrm{Optimize}\big(g_d^{(t)}, \Delta \mathcal{R}_d\big),
& \text{if } d \in \mathcal{D}_{\mathcal{G}^{(t)}}, \\
\mathrm{Generate}\big(\Delta \mathcal{R}_d\big),
& \text{otherwise}.
\end{cases}
\end{equation}
Here, $\mathcal{D}_{\mathcal{G}^{(t)}}$ denotes the set of diagnoses currently covered by the skill library $\mathcal{G}^{(t)}$. After applying this update across training diagnoses, we obtain the evolved skill library $\mathcal{G}^{\mathrm{Evo}}$.

This process improves both skill quality and skill coverage: existing skills become better aligned with real patient cases, while diagnoses missing from the initial guideline-derived library can be newly added.

\subsection{Skill Execution}

During inference, given an unseen patient case $x$, the LLM first generates a ranked differential diagnosis set $\mathcal{C}(x) \subseteq \mathcal{D}$. For each candidate diagnosis $d_i \in \mathcal{C}(x)$, \textit{GuideSkill} retrieves the corresponding skill from the evolved skill library by matching either the disease name or the ICD-10 code:
\begin{equation}
g_{d_i}
\leftarrow
\mathrm{Retrieve}(\mathcal{G}^{\mathrm{Evo}}, d_i).
\end{equation}

Because different skills may require different clinical features, the LLM extracts the skill-specific inputs needed by each retrieved skill:
\begin{equation}
\phi_{d_i}(x) =
\mathrm{LLM}_{\mathrm{feat}}(x, g_{d_i}),
\end{equation}
where $\phi_{d_i}(x)$ denotes the subset of patient features required to execute $g_{d_i}$. The retrieved skill is then executed on these extracted features to produce a skill-based evidence score:
\begin{equation}
s_{\mathrm{skill}}(x,d_i) =
g_{d_i}\big(\phi_{d_i}(x)\big).
\end{equation}

Meanwhile, the LLM-generated differential diagnosis list provides a ranking over the candidate set. We convert this rank into an LLM ranking score:
\begin{equation}
s_{\mathrm{LLM}}(x,d_i) =
\frac{1}{\mathrm{rank}_i + 1},
\end{equation}
where $\mathrm{rank}_i$ is the zero-based rank of candidate $d_i$. The final fused score combines the LLM ranking score with the skill-based evidence score:
\begin{equation}
s_{\mathrm{fuse}}(x,d_i) =
\alpha s_{\mathrm{LLM}}(x,d_i)
+
(1-\alpha)s_{\mathrm{skill}}(x,d_i),
\end{equation}
where $\alpha \in [0,1]$ controls the relative contribution of the two signals. By default, we set $\alpha=0.5$. The final diagnosis is selected as:
\begin{equation}
\hat{d} =
\arg\max_{d_i \in \mathcal{C}(x)}
s_{\mathrm{fuse}}(x,d_i).
\end{equation}

Compared with direct LLM inference, this process grounds the final decision in executable clinical rules while preserving the broad diagnostic capability of the LLM.

\begin{figure}[t]
\centering
\includegraphics[width=0.5\textwidth]{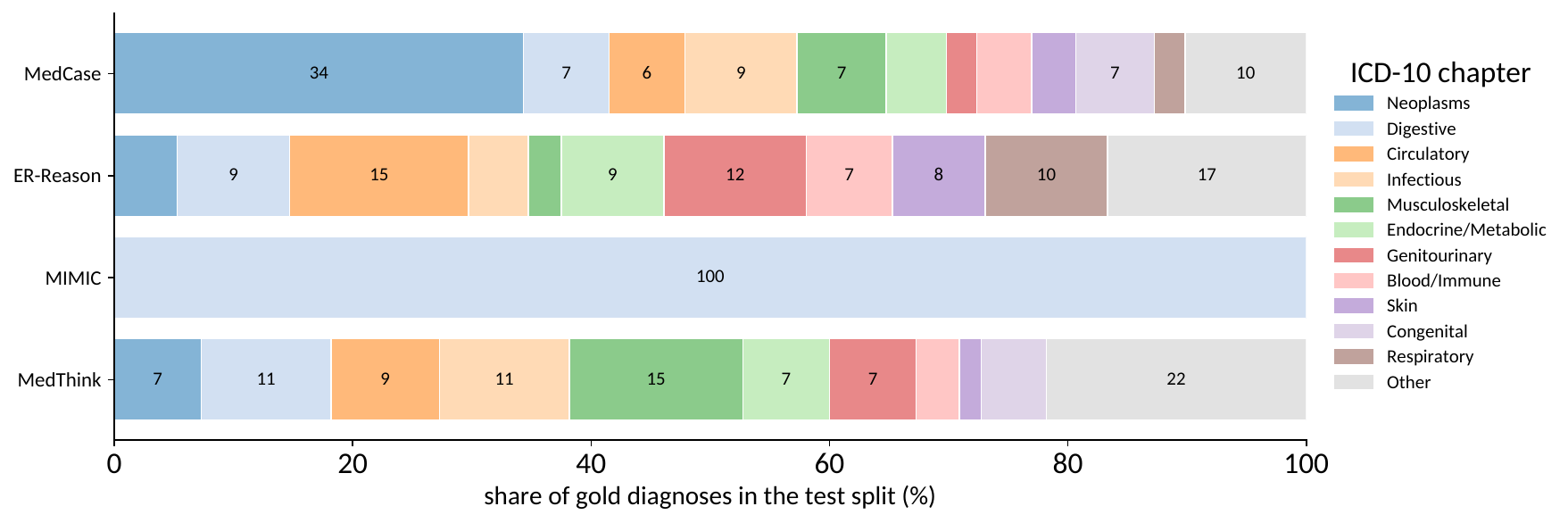}
\caption{Test-split diagnosis distribution by ICD-10 chapter across the four benchmarks. Each horizontal bar shows the percentage of gold diagnoses belonging to each chapter.}
\label{fig:icd_distribution}
\end{figure}

\begin{table}[t]
\centering
\resizebox{0.46\textwidth}{!}{%
\begin{tabular}{lccccc}
\toprule
\textbf{Statistic} & \textbf{MedCase} & \textbf{ER-Reason} & \textbf{MIMIC} & \textbf{MedThink} & \textbf{Total} \\
\midrule
\multicolumn{6}{l}{\textbf{Training split for evolution (\#)}} \\
\quad Cases & 11{,}598 & 1{,}235 & 219 & - & 13{,}052 \\
\quad Distinct ICD-10 categories & 463 & 232 & 5 & - & 473 \\
\quad New skills added & 263 & 102 & 0 & - & 267 \\
\midrule
\multicolumn{6}{l}{\textbf{Test split for inference (\#)}} \\
\quad Cases & 894 & 360 & 94 & 55 & 1{,}403 \\
\quad Distinct ICD-10 categories & 389 & 130 & 5 & 49 & 448 \\
\midrule
\multicolumn{6}{l}{\textbf{Skill coverage on test cases (\%)}} \\
\quad \textit{GuideSkill-Zero} & 43.1 & 79.2 & 100.0 & 50.9 & 56.5 \\
\quad \textit{GuideSkill-Evo} & \textbf{100.0} & \textbf{100.0} & \textbf{100.0} & \textbf{87.3} & \textbf{99.5 (\textcolor{langgreen}{+43.0})}\\
\bottomrule
\end{tabular}%
}
\caption{Benchmark statistics and skill coverage. \textit{GuideSkill-Evo} substantially improves test-case skill coverage over \textit{GuideSkill-Zero}, increasing overall coverage by 43.0 points.}
\label{tab:data_stats}
\end{table}

\begin{table*}[t]
\centering
\resizebox{1\textwidth}{!}{%
\begin{tabular}{cllccccc}
\toprule
 & \textbf{Base Model} & \textbf{Method} & \textbf{MedCaseReasoning} & \textbf{ER-Reason} & \textbf{MIMIC-CDM-FI} & \textbf{MedThink-Bench} & \textbf{Average} \\
\midrule
\multirow{8}{*}{\includegraphics[width=1.5em]{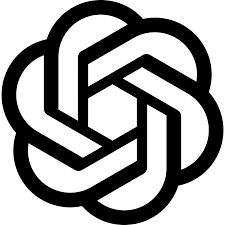}}
& \multirow{8}{*}{\textbf{GPT-5.4}}
& Direct & 25.73 & 42.22 & \textit{93.62} & 29.09 & 47.67 \\
& & CoT & 25.17 & 43.06 & 89.36 & \textbf{36.36} & 48.49 \\
& & 3-shot ICL & \textit{28.97} & \textit{43.89} & 92.55 & 34.55 & \textit{49.99} \\
& & RAG & 23.71 & 43.06 & \textit{93.62} & 32.73 & 48.28 \\
& & LLM DDx & 25.28 & 40.00 & 90.43 & 34.55 & 47.57 \\
& & LLM DDx + RAG & 25.39 & 39.72 & 90.43 & 27.27 & 45.70 \\
\cmidrule(lr){3-8}
& & \cellcolor{gray!10} \textcolor{black!55}{\textbf{\textit{GuideSkill-Zero}}} & \cellcolor{gray!10} \textcolor{black!55}{\textbf{\textit{35.01}}} & \cellcolor{gray!10} \textcolor{black!55}{\textbf{\textit{53.06}}} & \cellcolor{gray!10} \textcolor{black!55}{\textbf{\textit{95.74}}} & \cellcolor{gray!10} \textcolor{black!55}{\textbf{\textit{34.55}}} & \cellcolor{gray!10} \textcolor{black!55}{\textbf{\textit{54.59}}} \\
& & \cellcolor{langlightgreen!50} \textbf{GuideSkill-Evo} & \cellcolor{langlightgreen!50} \textbf{39.71}~\gain{+37.07\%} & \cellcolor{langlightgreen!50} \textbf{55.00}~\gain{+25.31\%} & \cellcolor{langlightgreen!50} \textbf{96.81}~\gain{+3.41\%} & \cellcolor{langlightgreen!50} \textbf{36.36}~\gain{+0.00\%} & \cellcolor{langlightgreen!50} \textbf{56.97}~\gain{+13.96\%} \\
\midrule
\multirow{8}{*}{\includegraphics[width=1.5em]{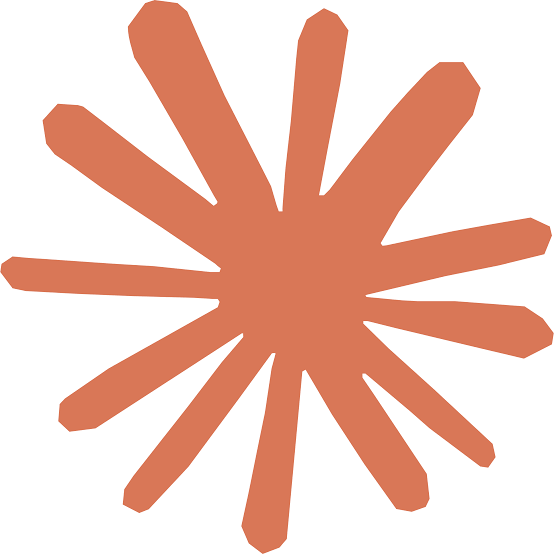}}
& \multirow{8}{*}{\textbf{Claude-Sonnet-4.6}}
& Direct & 23.27 & 41.11 & \textit{92.55} & \textit{34.55} & 47.87 \\
& & CoT & 23.94 & 39.17 & 89.36 & 32.73 & 46.30 \\
& & 3-shot ICL & \textit{25.95} & 39.94 & \textit{92.55} & \textit{34.55} & \textit{48.25} \\
& & RAG & 23.83 & \textit{43.06} & 91.49 & 29.09 & 46.87 \\
& & LLM DDx & 22.04 & 34.44 & \textit{92.55} & 30.91 & 44.99 \\
& & LLM DDx + RAG & 22.15 & 35.56 & 91.49 & 27.27 & 44.12 \\
\cmidrule(lr){3-8}
& & \cellcolor{gray!10} \textcolor{black!55}{\textbf{\textit{GuideSkill-Zero}}} & \cellcolor{gray!10} \textcolor{black!55}{\textbf{\textit{32.33}}} & \cellcolor{gray!10} \textcolor{black!55}{\textbf{\textit{47.22}}} & \cellcolor{gray!10} \textcolor{black!55}{\textbf{\textit{94.68}}} & \cellcolor{gray!10} \textcolor{black!55}{\textbf{\textit{25.45}}} & \cellcolor{gray!10} \textcolor{black!55}{\textbf{\textit{49.92}}} \\
& & \cellcolor{langlightgreen!50} \textbf{GuideSkill-Evo} & \cellcolor{langlightgreen!50} \textbf{40.27}~\gain{+55.18\%} & \cellcolor{langlightgreen!50} \textbf{50.56}~\gain{+17.42\%} & \cellcolor{langlightgreen!50} \textbf{95.74}~\gain{+3.45\%} & \cellcolor{langlightgreen!50} \textbf{40.00}~\gain{+15.77\%} & \cellcolor{langlightgreen!50} \textbf{56.64}~\gain{+17.39\%} \\
\midrule
\multirow{8}{*}{\includegraphics[width=1.5em]{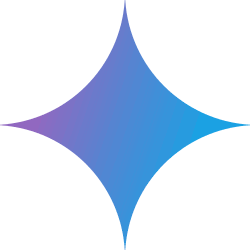}}
& \multirow{8}{*}{\textbf{MedGemma$_{\text{27B}}$}}
& Direct & 19.02 & 32.22 & 91.49 & 25.45 & 42.05 \\
& & CoT & 20.02 & 33.89 & \textit{92.55} & 25.45 & 42.98 \\
& & 3-shot ICL & \textit{20.25} & 30.28 & 91.49 & \textit{27.27} & 42.32 \\
& & RAG & 14.54 & 30.83 & 91.49 & 21.82 & 39.67 \\
& & LLM DDx & 18.57 & 39.17 & \textit{92.55} & 21.82 & \textit{43.03} \\
& & LLM DDx + RAG & 17.90 & \textit{39.44} & \textit{92.55} & 18.18 & 42.02 \\
\cmidrule(lr){3-8}
& & \cellcolor{gray!10} \textcolor{black!55}{\textbf{\textit{GuideSkill-Zero}}} & \cellcolor{gray!10} \textcolor{black!55}{\textbf{\textit{23.60}}} & \cellcolor{gray!10} \textcolor{black!55}{\textbf{\textit{44.72}}} & \cellcolor{gray!10} \textcolor{black!55}{\textbf{\textit{92.55}}} & \cellcolor{gray!10} \textcolor{black!55}{\textbf{\textit{27.27}}} & \cellcolor{gray!10} \textcolor{black!55}{\textbf{\textit{48.08}}} \\
& & \cellcolor{langlightgreen!50} \textbf{GuideSkill-Evo} & \cellcolor{langlightgreen!50} \textbf{25.95}~\gain{+28.15\%} & \cellcolor{langlightgreen!50} \textbf{48.89}~\gain{+23.96\%} & \cellcolor{langlightgreen!50} \textbf{93.62}~\gain{+1.16\%} & \cellcolor{langlightgreen!50} \textbf{30.91}~\gain{+13.35\%} & \cellcolor{langlightgreen!50} \textbf{49.84}~\gain{+15.83\%} \\
\midrule
\multirow{8}{*}{\includegraphics[width=1.5em]{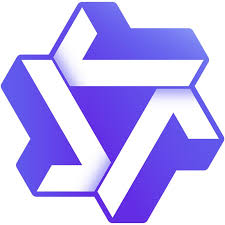}}
& \multirow{8}{*}{\textbf{Qwen3.5$_{\text{9B}}$}}
& Direct & 19.69 & 39.44 & 92.55 & 21.82 & 43.38 \\
& & CoT & 19.57 & 36.11 & \textit{93.62} & 23.64 & 43.24 \\
& & 3-shot ICL & \textit{20.69} & \textit{39.94} & 92.55 & 23.64 & \textit{44.21} \\
& & RAG & 16.55 & 33.33 & 90.43 & 23.64 & 40.99 \\
& & LLM DDx & 17.67 & 33.33 & \textit{93.62} & 20.00 & 41.16 \\
& & LLM DDx + RAG & 19.13 & 35.28 & 92.55 & 21.82 & 42.20 \\
\cmidrule(lr){3-8}
& & \cellcolor{gray!10} \textcolor{black!55}{\textbf{\textit{GuideSkill-Zero}}} & \cellcolor{gray!10} \textcolor{black!55}{\textbf{\textit{23.71}}} & \cellcolor{gray!10} \textcolor{black!55}{\textbf{\textit{41.67}}} & \cellcolor{gray!10} \textcolor{black!55}{\textbf{\textit{94.68}}} & \cellcolor{gray!10} \textcolor{black!55}{\textbf{\textit{25.45}}} & \cellcolor{gray!10} \textcolor{black!55}{\textbf{\textit{46.38}}} \\
& & \cellcolor{langlightgreen!50} \textbf{GuideSkill-Evo} & \cellcolor{langlightgreen!50} \textbf{26.17}~\gain{+26.49\%} & \cellcolor{langlightgreen!50} \textbf{46.39}~\gain{+16.15\%} & \cellcolor{langlightgreen!50} \textbf{96.81}~\gain{+3.41\%} & \cellcolor{langlightgreen!50} \textbf{34.55}~\gain{+46.15\%} & \cellcolor{langlightgreen!50} \textbf{50.98}~\gain{+15.31\%} \\
\bottomrule
\end{tabular}%
}
\caption{Main results across four benchmarks and four base models, reported as accuracy (\%). For each base model and column, \textbf{bold} indicates the best score, and \textit{italic} indicates the strongest baseline. \textcolor{langdarkgreen}{Green} numbers show the relative improvement of \textit{GuideSkill-Evo} over the strongest baseline. \textit{GuideSkill-Evo} achieves the best accuracy across all benchmarks and base models, with the largest gains over the strongest baselines and \textit{GuideSkill-Zero}.}
\label{tab:main_results}
\end{table*}

\section{Experiments}

\paragraph{Data.}

We evaluate \textit{GuideSkill-Zero} and \textit{GuideSkill-Evo} on four heterogeneous diagnostic-reasoning benchmarks: MedCaseReasoning~\cite{wu2025medcasereasoning}, ER-Reason~\cite{mehandru2025er}, MIMIC-CDM-FI~\cite{hager2024evaluation}, and MedThink-Bench~\cite{zhou2025automating}, spanning published case narratives, sequential emergency-department reasoning, full-information acute-abdominal diagnosis, and multistep medical QA. Training splits from the first three benchmarks are used for skill evolution, whereas MedThink-Bench is reserved for external evaluation and excluded from evolution. For cross-benchmark comparison, we normalize reference diagnoses to three-character ICD-10 categories, such as \textit{K35} for acute appendicitis, and evaluate category-level rather than subtype-level diagnosis. We use \textit{Claude-Sonnet-4.6} for skill initialization and evolution. The four test sets contain 1,403 cases across 448 ICD-10 categories (Table~\ref{tab:data_stats}), with their chapter-level diagnosis distributions shown in Figure~\ref{fig:icd_distribution}.

\paragraph{Baselines}
To assess \textit{GuideSkill} against representative inference-time alternatives under a common evaluation protocol, we compare six baselines: direct prompting, chain-of-thought prompting~\cite{wei2022cot}, 3-shot in-context learning~\cite{brown2020fewshot}, guideline RAG~\cite{lewis2020rag}, LLM-generated differential diagnosis (DDx), and LLM-generated DDx with guideline retrieval. These baselines test whether performance gains can be explained by explicit reasoning, case demonstrations, access to guideline text, broader candidate generation, or their combination. We evaluate every method using \textit{GPT-5.4}, \textit{Claude-Sonnet-4.6}, \textit{MedGemma-27B}, and \textit{Qwen3.5-9B}, covering proprietary and open-weight, general-purpose and medically specialized backbones.
We further compare \textit{GuideSkill} with representative parameter-update and structured-guideline alternatives on \textit{Qwen3.5-9B}: fine-tuning with guidelines~\cite{staniek2025training}, cases~\cite{wu2025medcasereasoning}, or both; RL with cases~\cite{chen2024huatuogpt}; guideline fine-tuning followed by case-based RL; and guidelines represented as decision trees~\cite{deng2026cpgprompt}. These methods span the principal supervision sources, optimization strategies, and guideline representations relevant to our setting. To ensure comparability, we implement them using the same backbone, data splits, evaluation protocol, and curated guideline corpus whenever applicable, rather than comparing with published results obtained on different tasks. The cited methods therefore motivate matched baseline adaptations rather than exact reproductions. For case-based training, we reserve 20\% of the training split for validation and use the remainder for optimization.

\paragraph{Evaluation.}
We report accuracy on each benchmark and the macro-average across the four benchmarks. Gold diagnoses are normalized to three-character WHO ICD-10 categories during preprocessing, and cases without a reliable mapping to a single diagnostic category are excluded. Because all methods are instructed to return an ICD-10 category code, we use exact equality between the predicted and reference codes as the primary correctness criterion; malformed or code-free outputs are counted as incorrect. Appendix~\ref{sec:exp_details} provides the model and inference settings, baseline implementations, training configurations, and complete evaluation protocol. Appendix~\ref{sec:benchmark_details} describes ICD-10 normalization, data filtering, and benchmark splits.


\begin{figure*}[t!]
\centering
\includegraphics[width=1\textwidth]{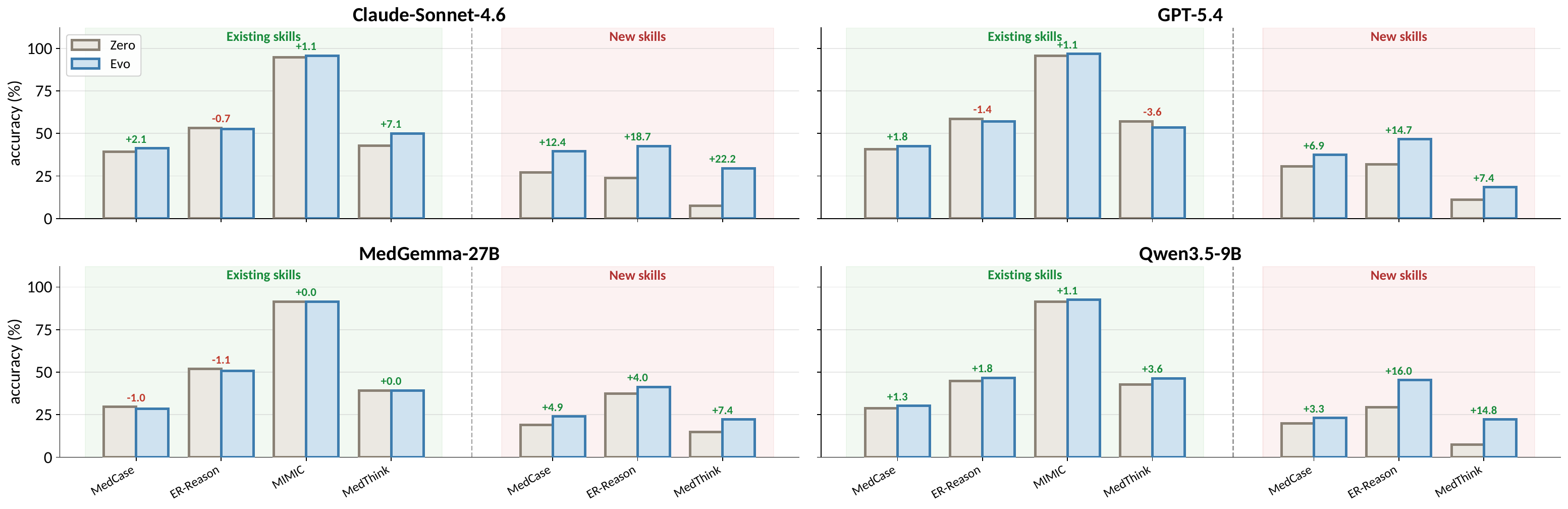}
\caption{Accuracy of \textit{GuideSkill-Zero} and \textit{GuideSkill-Evo} on initially covered and newly added ICD-10 categories. Evolution improves all new-skill settings while preserving or improving existing-skill performance in 11 of 16 settings.}
\label{fig:evo_vs_zero}
\end{figure*}

\begin{table*}[t]
\centering
\resizebox{0.97\linewidth}{!}{%
\begin{tabular}{lccccc}
\toprule
\textbf{Method} & \textbf{MedCaseReasoning} & \textbf{ER-Reason} & \textbf{MIMIC-CDM-FI} & \textbf{MedThink-Bench} & \textbf{Average} \\
\midrule
\raisebox{-0.5em}{\includegraphics[width=1.5em]{figures/icon/qwen.png}}~\textbf{Qwen3.5$_{\text{9B}}$} \\
\quad Direct Inference & 19.69 & 39.44 & 92.55 & 21.82 & 43.38 \\
\quad Fine-tuning w/ Guidelines \cite{staniek2025training} & 23.94 & 43.33 & 92.55 & 18.18 & 44.50 \\
\quad Fine-tuning w/ Cases \cite{wu2025medcasereasoning} & 23.27 & \textbf{46.94} & 92.55 & 18.18 & 45.24 \\
\quad Fine-tuning w/ Guidelines + Cases & 24.50 & 45.28 & \textit{93.62} & 16.36 & 44.94 \\
\quad RL w/ Cases \cite{chen2024huatuogpt} & \textit{25.95} & 42.50 & 92.55 & 16.36 & 44.34 \\
\quad Fine-tuning w/ Guidelines + RL w/ Cases & 24.94 & 43.06 & \textit{93.62} & 21.82 & \textit{45.86} \\
\quad Guidelines as Decision Trees \cite{deng2026cpgprompt} & 20.58 & 29.17 & 56.38 & \textit{32.73} & 34.72 \\
\midrule
\rowcolor{gray!10}
\quad \textcolor{black!55}{\textbf{\textit{GuideSkill-Zero (Ours)}}} & \textcolor{black!55}{\textbf{\textit{23.71}}} & \textcolor{black!55}{\textbf{\textit{41.67}}} & \textcolor{black!55}{\textbf{\textit{94.68}}} & \textcolor{black!55}{\textbf{\textit{25.45}}} & \textcolor{black!55}{\textbf{\textit{46.38}}} \\
\rowcolor{langlightgreen!50}
\quad \textbf{GuideSkill-Evo (Ours)} & \textbf{26.17} & \textit{46.39} & \textbf{96.81} & \textbf{34.55} & \textbf{50.98} \\
\bottomrule
\end{tabular}%
}
\caption{Training results on Qwen3.5-9B across four benchmarks, reported as accuracy (\%). For each column, \textbf{bold} indicates the best score. \textit{GuideSkill-Evo} achieves the best performance across all benchmarks and attains the highest average accuracy, consistently outperforming fine-tuning and RL-based training variants.}
\label{tab:training_results}
\end{table*}

\section{Results}

\paragraph{Main Results}
Table~\ref{tab:main_results} reports results across four backbone models. \textit{GuideSkill-Evo} achieves the highest macro-average accuracy for every backbone: 56.97 with \textit{GPT-5.4}, 56.64 with \textit{Claude-Sonnet-4.6}, 49.84 with \textit{MedGemma-27B}, and 50.98 with \textit{Qwen3.5-9B}. For \textit{GPT-5.4} and \textit{Claude-Sonnet-4.6}, these scores exceed the strongest non-\textit{GuideSkill} baseline by 6.98 and 8.39 percentage points, respectively. The gains therefore extend across proprietary, open-weight, general-purpose, and medically specialized backbones, indicating that the improvement is not tied to a particular LLM.

Even without case-based evolution, \textit{GuideSkill-Zero} often ranks second within each backbone block and outperforms guideline RAG for every backbone. This is notable because its guideline-derived library covers only a subset of the evaluated diagnoses; candidates without a matching skill rely primarily on the LLM ranking. Its competitive performance shows that guideline-derived executable skills provide substantial diagnostic value before using benchmark training cases. Rather than creating this benefit from scratch, evolution builds on it by expanding the library from 349 to 473 ICD-10 categories and increasing gold-label skill coverage from 56.5\% to 99.5\%. The resulting \textit{GuideSkill-Evo} improves over \textit{GuideSkill-Zero} on MedCaseReasoning, ER-Reason, and MedThink-Bench. On MedThink-Bench, which is excluded from evolution, it outperforms \textit{GuideSkill-Zero} across all four backbones, suggesting that case-derived updates can transfer to an unseen benchmark.

The advantage over guideline RAG further highlights the value of operationalizing guideline knowledge. Whereas RAG appends retrieved passages and relies on the LLM to interpret them, \textit{GuideSkill} executes disease-specific criteria to produce candidate-level support scores. To examine robustness, efficiency, and mechanism, additional analyses show that $\alpha=0.5$ is a competitive fusion setting and that accuracy largely saturates at $K=5$ (Figures~\ref{fig:alpha_sensitivity} and~\ref{fig:topk}). Batched feature grounding in \textit{GuideSkill-Effi} reduces estimated GPT-5.4 API cost by 72.7\% with a 0.34-point accuracy decrease on MedCaseReasoning (Appendix~\ref{sec:efficiency}; Table~\ref{tab:cost_efficiency}). A case study further illustrates how candidate-specific skill scores can correct an initially incorrect LLM ranking (Appendix~\ref{sec:case}).


\paragraph{Skill Evolution}
We analyze evolution from two complementary perspectives: skill coverage and downstream diagnostic utility. Gold-label skill coverage is the percentage of test cases whose reference ICD-10 category has a corresponding executable skill. As shown in Table~\ref{tab:data_stats}, evolution expands the library from 349 to 473 ICD-10 categories and increases coverage from 56.5\% with \textit{GuideSkill-Zero} to 99.5\% with \textit{GuideSkill-Evo}, filling nearly all gaps in the initial guideline-derived library.
Figure~\ref{fig:evo_vs_zero} examines whether this expansion improves newly covered diagnoses without degrading the original skills. For newly covered categories, \textit{GuideSkill-Evo} improves accuracy in all 12 available backbone--benchmark comparisons, with gains of 3.3--22.2 percentage points. For initially covered categories, accuracy improves or remains unchanged in 11 of 16 comparisons; gains reach 7.1 points, whereas declines are limited to 3.6 points. Evolution therefore substantially improves newly covered diagnoses while preserving or strengthening the initial skills in most settings.
To identify the remaining bottlenecks, we conduct an error analysis with \textit{Claude-Sonnet-4.6}. Of the 749 errors, 456 (60.9\%) occur because the reference ICD-10 category is absent from the candidate set and therefore cannot be recovered by downstream skill execution. Candidate recall is thus the dominant remaining failure mode; Appendix~\ref{sec:error_analysis} and Table~\ref{tab:error_analysis} provide the complete error breakdown.

\paragraph{Comparison with Training- and Guideline-Based Baselines}
Table~\ref{tab:training_results} compares \textit{GuideSkill} with parameter-update and structured-guideline baselines using the same \textit{Qwen3.5-9B} backbone. {GuideSkill-Evo} achieves the highest macro-average accuracy of 50.98, exceeding the strongest parameter-update baseline by 5.12 percentage points and Guidelines as Decision Trees by 16.26 points. It ranks first on three of four benchmarks; the only exception is ER-Reason, where case fine-tuning exceeds it by 0.55 points. Notably, \textit{GuideSkill-Zero} already achieves a macro-average of 46.38, surpassing the strongest parameter-update baseline before case-based skill evolution.
On MedThink-Bench, which is excluded from skill evolution and case-based parameter training, \textit{GuideSkill-Evo} achieves 34.55, compared with 21.82 for the strongest parameter-update baseline and 32.73 for Guidelines as Decision Trees. This indicates stronger transfer to an unseen benchmark. By storing diagnostic knowledge in external functions that directly execute disease-specific criteria, \textit{GuideSkill} improves accuracy without updating the backbone while keeping its decision logic explicit.


\begin{table}[t]
\centering
\resizebox{0.48\textwidth}{!}{%
\begin{tabular}{lcc}
\toprule
\textbf{Dataset} & \textbf{Executable Skill} & \textbf{Textual Skill} \\
\midrule
\raisebox{-0.3em}{\includegraphics[width=1.3em]{figures/icon/claude.png}}~\textbf{Claude-Sonnet-4.6} \\
\quad MedCaseReasoning & 45.00 \textcolor{langgreen}{\textbf{$\pm$ 0.00}} & 44.60 $\pm$ 0.80 \\
\quad ER-Reason & 63.00 \textcolor{langgreen}{\textbf{$\pm$ 0.00}} & 61.40 $\pm$ 0.80 \\
\quad MIMIC-CDM-FI & 95.74 \textcolor{langgreen}{\textbf{$\pm$ 0.00}} & 96.38 $\pm$ 0.52 \\
\quad MedThink-Bench & 40.00 \textcolor{langgreen}{\textbf{$\pm$ 0.00}} & 40.00 \textcolor{langgreen}{\textbf{$\pm$ 0.00}} \\
\midrule
\quad Average & 60.94 \textcolor{langgreen}{\textbf{$\pm$ 0.00}} & 60.60 $\pm$ 0.13 \\
\bottomrule
\end{tabular}%
}
\caption{Comparison between executable and textual skill representations using the same \textit{GuideSkill-Evo} candidate sets. Executable skills provide deterministic tier scoring after feature grounding, while textual skills rely on direct LLM interpretation of plain-text rubrics.}
\label{tab:text_vs_exec}
\end{table}

\paragraph{Comparison Between Executable and Textual Skills}
Under identical \textit{GuideSkill-Evo} candidate sets and fusion parameters, executable and textual skills achieve similar mean accuracy (60.94 versus 60.60), but only the executable variant shows no run-to-run accuracy variation over five evaluations, providing a more repeatable scoring interface (Table~\ref{tab:text_vs_exec}; Appendix~\ref{sec:text_vs_exec}).

\paragraph{Clinician Assessment of Skill Quality}
One clinician evaluated ten disease skills, five guideline-initialized and five case-derived, covering 42 initial rules, 100 evolution cases, 120 final rules, and 51 held-out cases. For each skill, the clinician first assigned an expected support tier (0--3) and confidence score to held-out cases without seeing the skill output, enabling comparison with the method-assigned tiers. For guideline-initialized skills, the clinician then assessed whether each initial rule was consistent with its cited guideline passage and whether clinically important criteria were omitted or incorrectly encoded. Next, the clinician rated whether each evolution case was suitable for skill evolution and whether it supported a new or revised rule. Finally, after reviewing the rule changes and final skill, the clinician recommended accepting, revising, or rejecting the skill. This staged protocol reduces anchoring from the skill output; Appendix~\ref{sec:questionnaire} provides the complete questionnaire.

As shown in Table~\ref{tab:clinician_eval}, 95.2\% of the initial rules were guideline-consistent, and all five guideline-initialized skills were judged to have no major clinical errors. Among evolution cases, 85.0\% were rated as suitable for skill evolution, with 64.7\% supporting a new or revised rule. On blinded held-out cases, all correct diagnoses received a support tier of at least 2, and 86.3\% of skill outputs were within one tier of the clinician judgment. The overall quadratic-weighted $\kappa$ was 0.60, indicating moderate clinician--skill agreement. Finally, 9/10 final skills were accepted unchanged or with only minor revisions. A detailed case review showed that the only final skill requiring major revision resulted from an evolution case that changed an originally correct rule into an inappropriate one by overgeneralizing a case-specific pattern rather than capturing a generalizable diagnostic rule. Overall, these results suggest that \textit{GuideSkill} produces clinically validated and largely acceptable skills through both skill initialization and evolution, while highlighting the need for careful validation of case-derived updates.


\begin{table}[t]
\centering
\small
\resizebox{\columnwidth}{!}{%
\begin{tabular}{lccc}
\toprule
\textbf{Measure} & \textbf{Guideline} & \textbf{Case-derived} & \textbf{Overall} \\
\midrule
\multicolumn{4}{l}{\textit{Initial rule and skill validity}} \\
Guideline-consistent rules & 95.2\% & -- & -- \\
Skills without major clinical errors & 5/5 & -- & -- \\
\midrule
\multicolumn{4}{l}{\textit{Evolution case utility}} \\
Cases suitable for skill evolution & 90.0\% & 80.0\% & 85.0\% \\
\quad\textit{supporting a new or revised rule} & \textit{60.0\%} & \textit{70.0\%} & \textit{64.7\%} \\
\midrule
\multicolumn{4}{l}{\textit{Blinded held-out case assessment}} \\
Correct diagnoses assigned tier $\geq$\,2 & 12/12 & 11/11 & 23/23 \\
Agreement within one tier & 88.0\% & 84.6\% & 86.3\% \\
Exact tier agreement & 52.0\% & 34.6\% & 43.1\% \\
Quadratic-weighted $\kappa$ & 0.64 & 0.55 & 0.60 \\
\midrule
\multicolumn{4}{l}{\textit{Final skill acceptability}} \\
Skills accepted unchanged/minor & 4/5 & 5/5 & 9/10 \\
\bottomrule
\end{tabular}%
}
\caption{Clinician assessment of guideline-initialized and case-derived skills. Tier-agreement metrics compare the clinician's blinded judgments with the corresponding skill outputs.}
\label{tab:clinician_eval}
\end{table}

\section{Conclusion}
We introduced \textit{GuideSkill}, an external reasoning layer that compiles guideline criteria into executable disease-specific skills and evolves them from labeled cases without updating the backbone. Across four benchmarks and four LLMs, guideline-initialized skills outperform guideline RAG, while evolution improves all 16 direct-inference comparisons and raises gold-label skill coverage from 56.5\% to 99.5\%. On \textit{Qwen3.5-9B}, \textit{GuideSkill-Evo} also exceeds the strongest matched parameter-update baseline. These results show that clinical knowledge can be maintained as a reusable, inspectable, and extensible execution layer across LLM backbones; broader clinician validation remains necessary before clinical use.

\clearpage

\bibliography{aaai2027}


\clearpage

\appendix
\section{Limitations}
\label{sec:limitations}

Although \textit{GuideSkill} substantially improves clinical reasoning accuracy by integrating executable clinical skills with LLM-based reasoning, it still has several limitations. First, the current skill library is built from a limited set of clinical guidelines. Many high-quality guidelines remain to be collected, curated, and incorporated, which may further improve the coverage and reliability of the skill library. Second, for scalability, the current framework performs skill initialization and skill evolution in a fully automated manner using LLMs. While this design makes \textit{GuideSkill} easy to scale, involving physicians in the verification process could further improve the quality and clinical validity of the generated skills, though at the cost of additional time and annotation effort. Finally, \textit{GuideSkill} is still a research prototype and is not intended for direct deployment in real clinical settings. Our goal is to provide an initial step toward combining executable clinical knowledge with LLM-based reasoning, and further clinical validation is required before practical use.

\begin{figure*}[t]
\centering
\includegraphics[width=\textwidth]{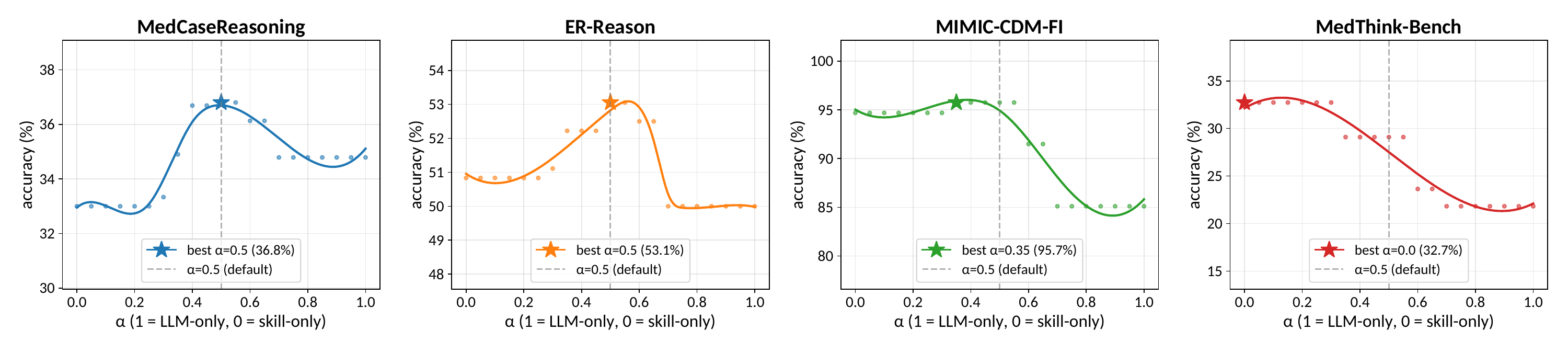}

\caption{Sensitivity analysis of the fusion weight $\alpha$ on four clinical reasoning benchmarks. $\alpha=1$ corresponds to LLM-only ranking, while $\alpha=0$ corresponds to skill-only scoring. The vertical dashed line marks the default setting $\alpha=0.5$. Across datasets, \textit{GuideSkill} remains relatively stable over a broad range of $\alpha$, showing that fusing LLM ranking with executable skill scores provides robust diagnostic performance.}
\label{fig:alpha_sensitivity}
\end{figure*}

\begin{figure}[t]
\centering
\includegraphics[width=0.4\textwidth]{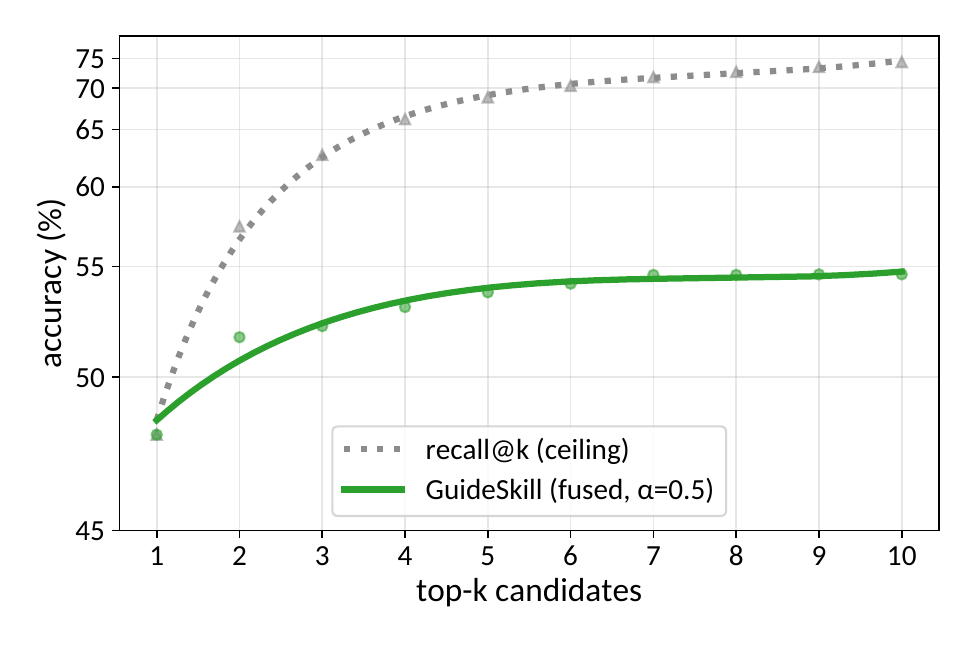}
\caption{
Sensitivity analysis of candidate set size. We vary the number of candidate diagnoses $K$ generated by the LLM and report the average accuracy. The dotted curve shows recall@$K$, which serves as an upper-bound ceiling. As $K$ increases, the final accuracy of \textit{GuideSkill} quickly saturates, indicating that a moderate candidate set provides a practical balance between candidate coverage and selection difficulty.
}
\label{fig:topk}
\end{figure}

\section{Preprocessing of Clinical Practice Guidelines}
\label{sec:guideline_preprocessing}

Our preprocessing pipeline serves two goals: (A) curating a corpus of \textit{usable} clinical practice guidelines from a large and heterogeneous source, and (B) compiling this free text into a library of \textit{executable} disease-level diagnostic skills.

\subsection{Guideline Corpus Curation}

We start from the publicly available \textit{epfl-llm/guidelines} corpus~\cite{chen2023meditron}, which is commonly used as a pretraining corpus for medical LLMs. The corpus contains $37{,}970$ clinical practice guideline (CPG) documents drawn from nine public sources. However, because these documents were scraped from online sources, the raw collection is highly noisy: some documents have missing or empty body text, some are extremely short or excessively long, and many contain content that is not directly relevant to actionable clinical guidance. In addition, the corpus varies substantially in length, ranging from $5$ to over $255{,}000$ words, with a median of $618$ words. We therefore apply a four-stage cascade filter to curate a usable guideline corpus.

\paragraph{Source filtering.}
We first retain only seven authoritative clinical guideline sources: NICE, PubMed, CMA, CDC, SPOR, WHO, and CCO. We discard the crowd-sourced and quality-inconsistent \textit{wikidoc} subset ($33{,}058$ documents), as well as the length-extreme \textit{icrc} subset ($49$ documents). This reduces the corpus from $37{,}970$ to $4{,}863$ documents.

\paragraph{Length filtering.}
We remove documents with missing body text or fewer than $100$ words, yielding $4{,}675$ documents.

\paragraph{Usability filtering.}
Even authoritative sources contain tables of contents, reference lists, methodology sections, disclaimers, announcements, and abstracts without actionable guidance. To remove such documents, we use an LLM-as-a-judge, \textit{Claude-Sonnet-4.6}, to determine whether each document, truncated to its first $12{,}000$ characters, contains \emph{actionable clinical guidance}, defined as concrete recommendations on diagnosis, treatment, screening, management, dosing, or eligibility. Only documents judged usable are retained, leaving $4{,}145$ documents.

\paragraph{Length-outlier truncation.}
Finally, we remove the longest $5\%$ of documents using a p$95$ word-count cutoff. This prevents overlong documents from dominating downstream context and diluting relevant content. The final curated corpus contains $\textbf{3{,}938}$ clinical practice guidelines, with a median length of $4{,}030$ words, a mean length of $5{,}046$ words, and a range of $101$–$18{,}662$ words. By source, it comprises $1{,}585$ NICE, $1{,}236$ PubMed, $377$ CMA, $374$ CDC, $184$ SPOR, $108$ WHO, and $74$ CCO documents.

\subsection{From Guidelines to Executable Diagnostic Skills}
Free-text guidelines cannot be executed directly. We therefore compile the curated corpus into a disease-indexed library of executable diagnostic skills through three steps.

\paragraph{Recommendation extraction.}
Using \textit{Claude-Opus-4.8}, we extract \textit{disease-diagnosis} recommendations from each guideline, namely rules of the form clinical findings or diagnostic criteria $\rightarrow$ confirm or rule out a specific disease $X$. Here, a recommendation refers to an actionable diagnostic rule that links observable patient evidence, such as symptoms, signs, laboratory results, imaging findings, or diagnostic criteria, to a disease-level conclusion. To ensure executability, we impose three strict constraints. First, we retain only rules whose output is a disease-level conclusion, explicitly excluding test-appropriateness statements, such as whether to order a scan, even when they mention diagnosis or staging. Second, each recommendation must map to a three-character ICD-10 category, such as \textit{K35}; recommendations that do not map to a valid disease category, such as non-disease findings codable only in the Z chapter, are discarded. Third, each recommendation is annotated with a disease name, its ICD-10 category, and an \texttt{is-us} flag indicating whether the content is specific to United States clinical practice. This step yields $1{,}200$ diagnostic recommendations spanning $349$ unique ICD-10 disease categories, of which $193$ are marked as US-specific.

\paragraph{Merging by disease.}
Because the same disease is often covered by multiple guidelines, we merge all recommendations within each ICD-10 category into a single self-contained, complementary, and de-duplicated diagnostic statement. During merging, we reconcile US-specific and non-US content using the \texttt{is-us} flag. This produces one merged diagnostic statement for each ICD-10 category, resulting in $349$ statements in total.

\paragraph{Skill synthesis and indexing.}
Each merged statement is compiled into an executable Python diagnostic function, or \textit{skill}, that takes structured features extracted from a patient case as input and returns a diagnostic confidence tier for the corresponding disease. The result is an ICD-10-indexed library of $349$ executable diagnostic skills. Overall, the pipeline distills $37{,}970$ heterogeneous documents into $3{,}938$ curated guidelines and further compiles them into $349$ disease-organized, deterministically executable skills, which serve as the knowledge base for the retrieval and fusion stages of our method.

\begin{table*}[t!]
\centering
\resizebox{0.7\linewidth}{!}{%
\begin{tabular}{lcccc}
\toprule
\textbf{Method} & \textbf{Accuracy} & \textbf{Input Tokens / Case} & \textbf{Output Tokens / Case} & \textbf{Cost / 10k Cases} \\
\midrule
\raisebox{-0.2em}{\includegraphics[width=1.1em]{figures/icon/gpt.png}}~\textbf{GPT-5.4} \\
\quad CoT & 25.17 & 332.87 & 1287.27 & 201.41 \\
\quad 3-shot ICL & 28.97 & 1178.87 & 1007.13 & 180.54 \\
\quad RAG & 23.71 & 771.70 & 841.33 & 145.49 \\
\quad LLM DDx & 25.28 & 775.10 & 1808.27 & 290.62 \\
\quad LLM DDx + RAG & 25.39 & 1244.03 & 1988.57 & 329.39 \\
\midrule
\rowcolor{gray!10}
\quad \textcolor{black!55}{\textbf{\textit{GuideSkill (Ours)}}} 
& \textcolor{black!55}{\textbf{\textit{39.71}}} 
& \textcolor{black!55}{\textbf{\textit{9722.60}}} 
& \textcolor{black!55}{\textbf{\textit{1106.30}}} 
& \textcolor{black!55}{\textbf{\textit{409.01}}} \\
\rowcolor{langlightgreen!50}
\quad \textbf{\textit{GuideSkill-Effi (Ours)}} 
& \textit{39.37} 
& 3207.70 
& \textbf{208.80} 
& \textbf{111.51} \\
\bottomrule
\end{tabular}%
}
\caption{Accuracy, token usage, and estimated inference cost of different prompting and reasoning methods. \textit{GuideSkill} achieves the highest accuracy, while \textit{GuideSkill-Effi} maintains nearly the same accuracy with substantially fewer output tokens and much lower inference cost.}
\label{tab:cost_efficiency}
\end{table*}

\section{Details of the Textual-Skill Comparison}
\label{sec:text_vs_exec}

We compare executable and textual representations using identical \textit{GuideSkill-Evo} candidate sets and the same fusion rule. Executable skills apply Python functions to LLM-grounded features to produce deterministic support tiers (0--3), whereas textual skills ask the LLM to assign tiers directly from equivalent plain-text rubrics. We evaluate 349 cases over five runs with \textit{Claude-Sonnet-4.6} at temperature 0 and report mean accuracy and run-to-run standard deviation.

\section{Sensitivity Analysis of Score Fusion}
\label{sec:alpha_sensitivity}

Figure 2 analyzes the sensitivity of \textit{GuideSkill} to the fusion weight $\alpha$ across the four benchmarks. Recall that $\alpha$ controls the relative contribution of the LLM ranking score and the executable skill score: $\alpha=1$ corresponds to relying only on the LLM ranking, while $\alpha=0$ corresponds to relying only on the skill score. In our main experiments, we use the default setting $\alpha=0.5$.

Overall, the fused scoring mechanism is robust across a broad range of $\alpha$ values, but the best setting varies slightly by benchmark. On MedCaseReasoning and ER-Reason, performance peaks around $\alpha=0.5$, indicating that both the LLM's ranking prior and the skill-based evidence contribute useful information. On MIMIC-CDM-FI, the best performance is obtained around $\alpha=0.35$, suggesting that the executable skills provide particularly strong diagnostic signal in this more structured setting. On MedThink-Bench, performance is highest near $\alpha=0$, indicating that skill scores are more reliable than the LLM ranking for these compact but challenging reasoning cases.

These results show that neither pure LLM ranking nor pure skill scoring is uniformly optimal across benchmarks. Instead, combining the two sources of evidence yields stable performance, with $\alpha=0.5$ serving as a simple default that performs competitively across datasets without benchmark-specific tuning.

\section{Effect of Top-K Candidate Diagnoses}
\label{sec:topk_analysis}
Figure 3 analyzes the effect of the candidate set size $K$ on diagnostic accuracy based on \textit{Claude-Sonnet-4.6}. In our main experiments, we set $K=5$ by default. As expected, recall@$K$ consistently increases as more candidate diagnoses are included, since a larger candidate set is more likely to contain the ground-truth diagnosis. However, the final accuracy of \textit{GuideSkill} does not always increase at the same rate, because the model must still select the correct diagnosis from a larger set of plausible candidates. On MedCaseReasoning and ER-Reason, \textit{GuideSkill} quickly reaches a stable performance after a small number of candidates, indicating that most useful diagnostic evidence is already captured in the top-ranked candidates. On MIMIC-CDM-FI, performance improves with larger $K$ and then saturates, closely following the high recall ceiling. On MedThink-Bench, increasing $K$ brings more noticeable gains, suggesting that challenging reasoning cases benefit from a broader candidate set. Overall, this analysis shows that $K=5$ provides a good trade-off between candidate coverage and decision complexity, and supports its use as the default setting.

\section{Efficiency Analysis}
\label{sec:efficiency}


We evaluate the efficiency of \textit{GuideSkill} against baseline methods to assess its potential for future deployment. The original \textit{GuideSkill} framework performs skill-level evidence extraction independently for each candidate skill, which enables accurate and fine-grained reasoning but incurs relatively high inference cost. To improve deployment efficiency, \textit{GuideSkill-Effi} performs feature extraction in a unified pass, allowing shared case-level evidence to be extracted once and reused across skills. Specifically, \textit{GuideSkill-Effi} is an efficiency-oriented variant that reduces the number of LLM calls per case from $1+N$ to exactly two, where $N$ is the number of matched skills. In \textit{GuideSkill}, inference consists of one differential-diagnosis proposal followed by one feature-extraction call for each matched skill. In \textit{GuideSkill-Effi}, the first call proposes the top-$K$ differential diagnosis as above, while the second call extracts the required features for \emph{all} matched skills in a single pass. The model is instructed to return only feature keys that are explicitly present in the case, with all other features defaulting to absent. Each skill is then executed locally without further LLM calls, and fusion proceeds identically to \textit{GuideSkill}. Because feature extraction is a form-filling task rather than a reasoning task, \textit{GuideSkill-Effi} runs both calls without reasoning.

This design preserves the main benefit of skill-guided reasoning while greatly reducing generation overhead. We evaluate both \textit{GuideSkill} and its efficient variant against baseline methods on the MedCaseReasoning benchmark, where both \textit{GuideSkill} variants use the evolved skill library. As shown in Table~3, \textit{GuideSkill-Effi} achieves 39.37\% accuracy, only 0.34 points lower than \textit{GuideSkill}, while reducing the estimated cost from 409.01 to 111.51 per 10k cases.

Importantly, \textit{GuideSkill-Effi} still substantially outperforms all baseline methods in accuracy, while achieving the lowest inference cost among all compared methods. This shows that the efficient design does not simply trade accuracy for lower cost, but provides a better accuracy-cost balance. This reduction is particularly meaningful because, in many commercial LLM APIs, output tokens are priced higher than input tokens. GPT-5.4 follows this common pricing pattern: the official API price is \$2.50 per 1M input tokens and \$15.00 per 1M output tokens, so output tokens are 6$\times$ more expensive than input tokens\footnote{https://developers.openai.com/api/docs/models/gpt-5.4}. As a result, inference cost is often dominated by output length. \textit{GuideSkill-Effi} directly addresses this bottleneck by transferring much of the reasoning process from free-form LLM generation to structured executable skills, yielding much shorter outputs with little accuracy degradation.

\section{Error Analysis of \textit{GuideSkill-Evo}}
\label{sec:error_analysis}

\begin{table}[t]
\centering
\resizebox{0.45\textwidth}{!}{%
\begin{tabular}{llc}
\toprule
\textbf{Stage} & \textbf{Clinical Type} & \textbf{\%} \\
\midrule
\multirow{3}{*}{Candidate omission (60.9\%)}
& Coding granularity & 74.8 \\
& Common diagnosis that should be listed & 14.3 \\
& Rare or atypical diagnosis & 11.0 \\
\midrule
\multirow{3}{*}{Skill under-scoring (12.6\%)}
& Adjacent same-system diagnosis & 40.4 \\
& Cross-system mimic & 36.2 \\
& Causal chain & 22.3 \\
\midrule
\multirow{3}{*}{Distractor selection (26.6\%)}
& Adjacent same-system diagnosis & 47.2 \\
& Causal chain & 27.6 \\
& Cross-system mimic & 24.6 \\
\bottomrule
\end{tabular}%
}
\caption{Error analysis of \textit{GuideSkill-Evo} with \textit{Claude-Sonnet-4.6} across all test cases. Stage percentages indicate each error stage's share among all analyzed errors; clinical-type percentages are computed within each stage.}
\label{tab:error_analysis}
\end{table}

Although \textit{GuideSkill-Evo} achieves the best overall performance, it still leaves substantial room for improvement. We conduct an error analysis on all test cases using \textit{Claude-Sonnet-4.6} as the backbone and categorize the remaining errors into three stages. The analysis is judge and cateogirze also by \textit{Claude-Sonnet-4.6} \textit{Recall-miss} errors occur when the gold diagnosis is not included in the LLM-generated candidate set. \textit{Skill-underscoring} errors occur when the gold diagnosis is recalled but its executable skill assigns an insufficient score. \textit{Lost-to-distractor} errors occur when the gold diagnosis is recalled and scored, but another plausible candidate receives a higher fused score.

As shown in Table~6, the largest source of error is recall failure: in 456 cases, the gold ICD-10 category is not included in the candidate set, making it impossible for skill execution to recover the correct answer. Most recall-miss errors are due to coding granularity, where the LLM proposes a clinically related diagnosis but not the exact ICD-10 category required by evaluation. Among cases where the gold diagnosis is recalled, errors often arise from fine-grained clinical distinctions. Skill-underscoring errors are dominated by adjacent same-system diagnoses and cross-system mimics, suggesting that some skills still underweight discriminative evidence. Lost-to-distractor errors show a similar pattern: the correct diagnosis is present, but a nearby diagnosis, causal downstream condition, or cross-system mimic receives a stronger fused score. These results suggest that future improvements should target both candidate recall at the ICD-10 category level and finer-grained skill calibration among clinically similar diagnoses.


\section{Experiment Details}
\label{sec:exp_details}

\paragraph{Models.}
We evaluate all methods using four backbone LLMs spanning both proprietary and open-weight models. The proprietary models are accessed through Microsoft Azure\footnote{https://azure.microsoft.com/en-us}: Claude-Sonnet-4.6 (\texttt{claude-sonnet-4-6})~\cite{anthropic2026sonnet46} and GPT-5.4 (\texttt{gpt-5.4})~\cite{openai2026gpt54}. The open-weight models are served locally with vLLM~\cite{kwon2023pagedattention}: MedGemma-27B (\texttt{google/medgemma-27b-it})~\cite{google2026medgemma} and Qwen3.5-9B (\texttt{Qwen/Qwen3.5-9B})~\cite{qwen2026qwen35}. We use Claude-Sonnet-4.6 for skill initialization and skill evolution in \textit{GuideSkill}. For experiments involving non-public clinical data from PhysioNet, we follow the PhysioNet responsible-use guidance for MIMIC data with LLMs\footnote{https://physionet.org/news/post/llm-responsible-use/}. Specifically, restricted clinical data are used with proprietary models only through an institutionally approved Azure deployment that ensures zero data retention, no training on submitted data, and no human review of prompts or outputs. The locally served open-weight models provide a fully controlled deployment path.

\paragraph{LLM Inference.}
Proprietary models are queried through the Azure API. For Claude backbones, we set temperature to $0$ and do not enable extended thinking. GPT-5.4 is queried with the API default settings, including the default temperature of $1$ and the default reasoning configuration. Open-weight models are served through vLLM with a $32{,}768$-token context window. For Qwen3.5-9B, we set temperature to $0$ and explicitly disable thinking mode by setting \texttt{enable\_thinking=False}, so that the final answer is returned directly rather than embedded in a long reasoning trace. For MedGemma-27B, we use temperature $0$; since it is not a reasoning model, no thinking mode is used. Unless otherwise noted, generation is capped at $4{,}096$ tokens.

\paragraph{Skill Evolution.}
We evolve the skill library using a merged training set from MedCaseReasoning, ER-Reason, and MIMIC-CDM-FI, with no overlap with the test splits. Each case is labeled with its three-character ICD-10 category. The evolution pipeline mirrors skill initialization and also uses \textit{claude-sonnet-4-6}. First, in \emph{rationale generation}, the model is given each training case and its gold diagnosis, and is asked to extract the key clinical variables and explain why they support that diagnosis. Second, in \emph{case-to-recommendation}, cases are grouped by ICD-10 label and distilled into a single recommendation that captures only criteria recurring across cases, while discarding case-specific incidentals. When a matching guideline recommendation already exists, it is used as the backbone and is never weakened. We process cases in batches of $10$ using a map-reduce strategy: the model first drafts recommendations for each batch and then iteratively refines them, preventing prompt overflow and reducing specificity loss. Third, in \emph{recommendation-to-skill}, the resulting recommendations are compiled using the same generator and tier contract as in skill initialization. Skill evolution yields $473$ case-derived skills. After merging them with the initialized skill library, we obtain the evolved skill library covering $473$ ICD-10 categories, including $206$ categories shared with the initialized library and $267$ newly added categories from training cases.

\paragraph{Baselines.}
We compare \textit{GuideSkill} with both prompting-based and training-based baselines, using the same backbone models whenever applicable. For prompting-based baselines, we include: (i) \textbf{\textit{Direct}}, which directly prompts the model to produce the final diagnosis; (ii) \textbf{\textit{CoT}}, which performs chain-of-thought diagnostic reasoning before producing the final answer~\cite{wei2022cot}; (iii) \textbf{\textit{3-shot ICL}}, which uses three fixed in-domain demonstrations from each dataset's training split~\cite{brown2020fewshot}. Since MedThink-Bench does not provide a training split, we use demonstrations from MedCaseReasoning; (iv) \textbf{\textit{RAG}}, which retrieves from curated guideline recommendations and prepends the top-$5$ passages retrieved by \textit{text-embedding-3-small}~\cite{openai2026embedding} as context before answering~\cite{lewis2020rag}; (v) \textbf{\textit{LLM DDx}}, a two-pass differential-diagnosis baseline that first proposes the top-$5$ candidate diagnoses and then selects one final answer without using external knowledge; and (vi) \textbf{\textit{LLM DDx + RAG}}, which extends the two-pass differential-diagnosis baseline by conditioning the final selection on the same retrieved guideline context used in RAG.

For training-based baselines, we conduct comparisons on \textit{Qwen-3.5} and include: (vii) \textbf{\textit{Fine-tuning w/ Guidelines}}, which fine-tunes the model on all curated guideline recommendations; (viii) \textbf{\textit{Fine-tuning w/ Cases}}, which fine-tunes the model on training cases with gold diagnoses; (ix) \textbf{\textit{Fine-tuning w/ Guidelines + Cases}}, which fine-tunes on the union of guideline-derived supervision and case-level supervision; (x) \textbf{\textit{RL w/ Cases}}, which applies reinforcement learning on training cases using diagnosis correctness as the reward signal; and (xi) \textbf{\textit{Fine-tuning w/ Guidelines + RL w/ Cases}}, which first fine-tunes the model with guideline supervision and then further optimizes it with case-level reinforcement learning. All training baselines are implemented with verl\footnote{https://github.com/verl-project/verl} using the default training configuration. We also include (xii) \textbf{\textit{Guidelines as Decision Trees}}, which converts guideline recommendations into decision-tree-style reasoning structures and uses them as explicit diagnostic guidance. We use the official implementation released by the original paper \cite{deng2026cpgprompt}.

\paragraph{Details of Training Baselines.}
For all trainable variants, we apply LoRA~\cite{hu2021lora} to all linear layers with rank 16 and $\alpha=32$. Training uses bfloat16 precision, gradient checkpointing, AdamW optimization~\cite{loshchilov2019adamw}, a learning rate of $1\times10^{-5}$, a constant learning-rate schedule with 10 warmup steps, a global batch size of 512, a maximum sequence length of 8,192 tokens, and three training epochs. For case-based learning, we randomly split the original training set into training and validation subsets using an 8:2 ratio. Guideline SFT is performed on the clinical guideline corpus, whereas case SFT is conducted on standardized diagnosis cases using answer-only supervision. The \textit{Guidelines + Cases} setting performs these two stages sequentially. For RL, we adopt GRPO~\cite{shao2024deepseekmath} with a learning rate of $5\times10^{-6}$, two training epochs, a batch size of 32, and 24 rollouts per prompt. The maximum prompt and response lengths are set to 8,192 and 512 tokens, respectively. We use one policy-update epoch, a KL coefficient of 0.005, an entropy coefficient of 0, a rollout temperature of 1.0, and top-$p$ of 1.0. Checkpoints are saved after every SFT epoch and every 10 RL updates, and the best checkpoint is selected based on validation performance. All experiments are conducted on a single server equipped with eight NVIDIA RTX 5090 GPUs.

\paragraph{Evaluation Protocol.}
We evaluate predictions at the three-character ICD-10 category level: a prediction is counted as correct if it maps to the same ICD-10 category as the gold diagnosis. Semantic matching between the predicted diagnosis and the gold label is performed using an LLM-as-a-judge, \textit{claude-haiku-4-5}, with temperature set to $0$. The judge treats synonyms, abbreviations, subtype-level differences, and clinically equivalent paraphrases as matches when they fall within the same ICD-10 category. LLM judges can exhibit systematic biases and reasoning limitations~\cite{zheng2023judging}; we therefore identify judge-based semantic matching as a limitation of the evaluation. The full LLM-as-a-judge prompt is shown below.

\begin{tcolorbox}[
    colback=gray!5,
    colframe=langgreen,
    title=\textbf{\small LLM Judge Prompt for ICD-10 Category Matching},
    listing only,
    listing options={
        basicstyle=\ttfamily\small,
        breaklines=true,
        breakatwhitespace=true,
        columns=fullflexible
    }
]
Decide whether the predicted diagnosis and the ground-truth diagnosis refer to the same disease at the ICD-10 3-character category level (e.g., "K35").

Judge at the category granularity: they match if they fall under the same ICD-10 3-character category, even if the subtype or wording differs.

Predicted diagnosis:
\{pred\}

Ground-truth diagnosis:
\{gold\}

Please answer directly with "yes" or "no".
\end{tcolorbox}

\section{Benchmark Details and Examples}
\label{sec:benchmark_details}

The four benchmarks originate from different data sources and represent case reports, emergency-department records, structured clinical decision-making cases, and medical QA vignettes. Training splits from MedCaseReasoning, ER-Reason, and MIMIC-CDM-FI are used for skill evolution and parameter-update baselines; the corresponding held-out test splits and the MedThink-Bench test set are used for evaluation. The benchmark data are separate from the guideline corpus used to initialize \textit{GuideSkill-Zero}.

\begin{itemize}[leftmargin=*, itemsep=0pt, labelsep=5pt, topsep=0pt]
    \item \textbf{MedCaseReasoning} \cite{wu2025medcasereasoning}: A long-form open diagnostic reasoning benchmark based on open-access clinical case reports from the New England Journal of Medicine Clinicopathological Conferences (NEJM CPC). It evaluates case-based differential diagnosis over broad and long-tailed clinical conditions.

\begin{tcolorbox}[
    colback=gray!5,
    colframe=langgreen!80,
    title=\textbf{Example: MedCaseReasoning},
    listing only,
    listing options={
        basicstyle=\ttfamily\small,
        breaklines=true,
        breakatwhitespace=true,
        columns=fullflexible
    }
]
Case: A 30-year-old man presented with several months of pain, tenderness, and swelling on the left side of his palate. T2-weighted MRI showed a hyperintense lesion, and biopsy suggested a soft-tissue tumor. Given the case, what is the final diagnosis? [case report excerpt; approximately 890 words]

Diagnosis: Other disorders of nerve roots and plexuses.
ICD-10: G54.
\end{tcolorbox}

    \item \textbf{ER-Reason} \cite{mehandru2025er}: An emergency-department diagnosis prediction benchmark derived from real emergency-room patient records. It evaluates whether a model can infer the final diagnosis from noisy, time-sensitive clinical presentations in the emergency setting.

\begin{tcolorbox}[
    colback=gray!5,
    colframe=langgreen!80,
    title=\textbf{Example: ER-Reason},
    listing only,
    listing options={
        basicstyle=\ttfamily\small,
        breaklines=true,
        breakatwhitespace=true,
        columns=fullflexible
    }
]
Case: Age: 26; Sex: Female; Chief Complaint: hematuria. The record includes a full emergency-department note with pregnancy and delivery history, normal spontaneous vaginal delivery, second-degree perineal laceration repair, hematuria with dysuria, lower abdominal pain, and sexual history. [long EHR excerpt; approximately 15,400 words]

Diagnosis: Tubulo-interstitial nephritis, not specified as acute or chronic.
ICD-10: N12.
\end{tcolorbox}

    \item \textbf{MIMIC-CDM-FI} \cite{hager2024evaluation}: A full-information open clinical decision-making benchmark derived from MIMIC-IV, which is based on electronic health records from Beth Israel Deaconess Medical Center. It evaluates full-information clinical decision making.

\begin{tcolorbox}[
    colback=gray!5,
    colframe=langgreen!80,
    title=\textbf{Example: MIMIC-CDM-FI},
    listing only,
    listing options={
        basicstyle=\ttfamily\small,
        breaklines=true,
        breakatwhitespace=true,
        columns=fullflexible
    }
]
Case: Patient history describes 10 hours of abdominal pain that began near the umbilicus and migrated to the right lower quadrant, with one episode of diarrhea and no nausea, vomiting, or fever. Physical examination shows right-lower-quadrant tenderness, positive obturator sign, and negative Rovsing sign. Laboratory tests include urea nitrogen, sodium, potassium, lipase, creatinine, and other values. [structured history, exam, and laboratory excerpt; approximately 4,000 words]

Diagnosis: Acute appendicitis.
ICD-10: K35.
\end{tcolorbox}

    \item \textbf{MedThink-Bench}~\citep{zhou2025automating}: An expert-curated medical reasoning benchmark constructed from ten publicly available medical QA datasets. The benchmark filters for complex questions requiring multi-step reasoning across ten medical domains. In our setting, we use it as a compact but challenging open-ended diagnosis-oriented reasoning benchmark.

\begin{tcolorbox}[
    colback=gray!5,
    colframe=langgreen!80,
    title=\textbf{Example: MedThink-Bench},
    listing only,
    listing options={
        basicstyle=\ttfamily\small,
        breaklines=true,
        breakatwhitespace=true,
        columns=fullflexible
    }
]
Case: A 23-year-old female weightlifter presents with neck and right shoulder pain. Osteopathic structural examination reveals restricted motion of Sibson's fascia, and the clavicles appear asymmetric. What is the most likely diagnosis? [medical QA excerpt; approximately 780 words]

Diagnosis: Other acquired deformities of musculoskeletal system.
ICD-10: M95.
\end{tcolorbox}
\end{itemize}

\paragraph{Preprocessing and Data Split.}
To normalize answer labels across benchmarks, we first map all gold diagnoses to ICD-10 categories. Specifically, we use \textit{Claude-Sonnet-4.6} to convert each dataset's original answer label into an ICD-10 code. We discard examples for which no reliable ICD-10 mapping can be obtained. We also remove examples whose gold answer does not correspond to a single disease diagnosis, since our evaluation requires one normalized diagnostic target per case.

After preprocessing, we randomly split MedCaseReasoning, ER-Reason, and MIMIC-CDM-FI into training and test sets. MedThink-Bench does not provide a training split in our setting, so we use it only for evaluation. The final processed data contain 11,598 training and 894 test cases for MedCaseReasoning, 1,235 training and 360 test cases for ER-Reason, 219 training and 94 test cases for MIMIC-CDM-FI, and 55 test cases for MedThink-Bench. In total, our experiments use 13,052 training cases and 1,403 test cases after ICD-10 normalization and filtering.

\section{Clinician Assessment Questionnaire}
\label{sec:questionnaire}

This appendix gives the complete instrument used for the clinician assessment reported in
Table~\ref{tab:clinician_eval}, together with the order in which material was presented. The
study was delivered as a single-page web application; each packet covers one disease skill, and
the clinician worked through the screens in a fixed order without being able to see later
material in advance.

\subsection{Presentation Order}

Each packet is presented as four screens. Critically, the \emph{held-out screen comes first}:
the clinician assigns an expected support tier to unseen cases before any part of the skill ---
guideline passages, rule tables, or code --- is revealed. Presenting the skill first would let
its output anchor the expected tiers and inflate agreement.

\begin{enumerate}[leftmargin=*, itemsep=1pt, topsep=2pt]
  \item \textbf{Held-out cases} (Q5). Blinded tier assignment on cases never used to build the
        skill. The skill's own output stays hidden throughout.
  \item \textbf{Guideline and initial skill} (Q1, Q2). Numbered guideline passages
        \textsf{G1--G$n$}, a human-readable rule table \textsf{R1--R$n$} linking each rule to
        its supporting passage, and the initial Python function in an expandable panel.
  \item \textbf{Evolution cases} (Q3). Ten cases per skill, unlabeled and in fixed order, with
        the final skill still hidden.
  \item \textbf{Initial-to-final update} (Q4). The change log with per-change provenance, the
        final rule table, and the final Python function; the disposition is recorded last.
\end{enumerate}

For \emph{case-derived} skills there is no guideline-initialized predecessor, so Screen~2 and
questions Q1--Q2 are omitted (three screens, Q3--Q5). These packets are explicitly labeled
``case-derived skill; no guideline-derived initialization.'' In both tracks, ``cannot assess /
outside expertise'' is available on every question and is excluded from the corresponding
denominator rather than treated as a negative rating.

\subsection{Questions}

\paragraph{Q1 --- Guideline support (per rule; guideline-initialized skills only).}
\emph{For each rule in the initial skill, how well is it supported by the provided guideline
passages?}
\begin{itemize}[leftmargin=*, itemsep=0pt, topsep=2pt]
  \item Fully supported
  \item Mostly supported (minor interpretation)
  \item Partially supported (substantial interpretation)
  \item Unsupported or contradictory
  \item Cannot assess
\end{itemize}
Table~\ref{tab:clinician_eval} reports two thresholds on this scale: rules rated \emph{fully or
mostly supported}, and rules \emph{not} rated unsupported or contradictory.

\paragraph{Q2 --- Errors or omissions in the initial skill (per skill).}
\emph{Based only on the provided guideline passages, does the initial skill omit or incorrectly
encode any clinically important criterion, threshold, exclusion, exception, or logical
relationship?}
\begin{itemize}[leftmargin=*, itemsep=0pt, topsep=2pt]
  \item No clinically important problem
  \item Minor problem (unlikely to change the support tier)
  \item Major problem (could change the support tier)
  \item Potentially dangerous / seriously misleading
  \item Cannot assess
\end{itemize}
Q1 and Q2 are locked before the clinician proceeds to the evolution cases.

\paragraph{Q3 --- Contribution of an evolution case (per case).}
\emph{If a rule for this skill were written from this case, what would it do to future
patients?}
\begin{itemize}[leftmargin=*, itemsep=0pt, topsep=2pt]
  \item[A] Add or change a rule --- shows a criterion the skill needs and might otherwise miss
  \item[B] Keep the rules as they are --- textbook presentation; confirms what a skill would
        already check
  \item[C] Would apply to almost no one else --- a rule from this case would rarely fire again;
        harmless but useless
  \item[D] Would fire on the wrong patients --- a rule from this case would misjudge future
        patients (scores a mimic as this disease, drops a needed requirement, or relies on a
        non-specific feature)
  \item[E] Cannot assess
\end{itemize}
The deciding test between C and D is stated in the instrument: \emph{would a rule taken from
this case ever fire on a patient who does not have this disease? No~$\rightarrow$~C;
yes~$\rightarrow$~D.} An optional free-text field records the finding or omission that drove the
rating. Cases rated A, B, or C are counted as generalizable in
Table~\ref{tab:clinician_eval}; D marks a case that should not influence the skill.

\paragraph{Q4 --- Final update and disposition (per skill).}
\emph{After reviewing the guideline, the cases, and the initial-to-final diff, what is your
recommendation for the final skill?}
\begin{itemize}[leftmargin=*, itemsep=0pt, topsep=2pt]
  \item Accept unchanged
  \item Accept with minor edits
  \item Major revision required
  \item Reject
  \item Cannot assess
\end{itemize}
If the recommendation is not ``accept unchanged'', the clinician selects all applicable concerns
from: insufficient case support; case- or dataset-specific pattern; conflicts with the
guideline; weakens a guideline-supported requirement; omits an important criterion; treats an
unreported feature as absent; treats an unexcluded mimic as excluded; incorrect threshold, unit,
negation, or AND/OR logic; inappropriate support tier; other. A free-text field records the
single most important required change.

\paragraph{Q5 --- Expected support tier (per held-out case).}
\emph{Based on the available evidence, what diagnostic-support tier should a skill for this
disease assign to this case?}
\begin{itemize}[leftmargin=*, itemsep=0pt, topsep=2pt]
  \item 3 --- Confirmed / highest support
  \item 2 --- Strongly suggestive
  \item 1 --- Compatible
  \item 0 --- Not supported
  \item Insufficient information to assign a tier
  \item Cannot assess
\end{itemize}
A confidence rating (low / moderate / high) accompanies each tier. These are the same four tiers
the skills themselves emit, so the clinician's blinded judgment and the skill's executed tier
are directly comparable; agreement is computed only after all held-out cases for a packet are
submitted.

\subsection{Held-Out Case Construction}

Held-out cases are drawn from the MedCaseReasoning test split and are disjoint from every case
used to build the skill. Each packet mixes two kinds, and the kind is \emph{not} shown to the
clinician:

\begin{itemize}[leftmargin=*, itemsep=1pt, topsep=2pt]
  \item \textbf{Correct diagnoses} (23 cases): the reviewed category is the gold diagnosis.
  \item \textbf{Look-alikes} (28 cases): the gold diagnosis is a different category, but the
        reviewed category appeared in the model's proposed differential --- an operational,
        reproducible definition of clinical similarity.
\end{itemize}

Because both kinds are presented identically, Q5 measures whether the skill's tier tracks
clinical judgment on cases that do and do not warrant support, rather than only on positives.

\clearpage

\section{Case Study}
\label{sec:case}

We present a representative MedCaseReasoning example as a case study of \textit{GuideSkill}. In this example, \textit{GuideSkill} corrects the base LLM’s plausible but incorrect top-ranked diagnosis. The patient is a 60-year-old woman with dyspepsia, unintentional weight loss, postprandial vomiting, gastric ulcers on endoscopy, and biopsy showing large confluent non-caseating epithelioid granulomas. The base LLM initially ranks Crohn's disease as the most likely diagnosis, likely because gastrointestinal ulcers and granulomas are common cues for Crohn's disease. However, after executing candidate-specific skills, \textit{GuideSkill} assigns different diagnostic strengths to the same evidence. The Crohn's disease skill treats granulomas as a specific but insufficient feature and stops at tier 2, while the sarcoidosis skill treats non-caseating granulomas as confirmatory histologic evidence and assigns tier 3. The fusion step therefore overturns the LLM ranking and selects sarcoidosis, matching the gold diagnosis.

The key mechanism is that the two executed skills assign different evidential status to the same grounded finding. For sarcoidosis, non-caseating granulomas with unsupported mimics enter a direct tier-3 branch. For Crohn's disease, granulomas alone are only a specific tier-2 feature unless additional Crohn's-specific criteria or mimic-exclusion requirements are satisfied. Thus, the same clinical evidence is sufficient to confirm sarcoidosis but only suggestive for Crohn's disease, making the tier-based fusion interpretable.

\clearpage

\begin{table*}[t]
\centering
\small
\begin{tabular}{p{0.18\textwidth} p{0.76\textwidth}}
\toprule
\textbf{Field} & \textbf{Content} \\
\midrule
Case & A 60-year-old woman with no medical comorbidities presented with a 2-month history of dyspepsia, unintentional weight loss, and anorexia. She noted postprandial fullness, early satiety, and multiple episodes of non-bilious, non-projectile vomiting occurring 20--30 minutes after meals, containing undigested food. She had no fever, abdominal pain, gastrointestinal bleeding, or history of medication use. On examination, she was dehydrated and tachycardic. Abdominal examination revealed a distended and tender upper abdomen; the liver edge was palpable 2 cm below the right costal margin. Laboratory tests showed a low hemoglobin level with otherwise normal routine studies. Upper endoscopy demonstrated an oval pre-pyloric ulcer with erythematous, everted margins and a whitish exudate at the base, surrounded by normal mucosa, and an irregular fundal ulcer with inverted margins and whitish exudate, surrounded by normal mucosa. Multiple gastric biopsies revealed patchy chronic inflammation, mild crypt architectural disarray, and several large confluent non-caseating epithelioid granulomas. \\
\midrule
Ground Truth Diagnosis & Sarcoidosis (D86) \\
\midrule
LLM Top-5 Differential Diagnosis &
\begin{tabular}[t]{@{}ll@{}}
(1) & Crohn's disease [regional enteritis] (K50) \\
(2) & Sarcoidosis (D86) \\
(3) & Malignant neoplasm of stomach (C16) \\
(4) & Tuberculosis of other organs (A18) \\
(5) & Gastric ulcer (K25)
\end{tabular}
\\
\midrule
Skill Execution Trace &
\begin{tabular}[t]{@{}p{0.25\textwidth} p{0.11\textwidth} p{0.36\textwidth}@{}}
Crohn's disease (K50) & tier 2 / 0.667 & Granulomas and GI ulcers are specific evidence, but the case lacks decisive Crohn's evidence such as transmural or terminal ileal involvement, and direct-confirmation branches require mimics to be excluded. \\
Sarcoidosis (D86) & tier 3 / 1.000 & Non-caseating epithelioid granulomas provide confirmatory histologic evidence for sarcoidosis when competing mimics are not supported. \\
Malignant neoplasm of stomach (C16) & tier 0 / 0.000 & Endoscopic ulcers raise concern, but biopsy does not report malignant cells. \\
Tuberculosis of other organs (A18) & tier 1 / 0.333 & Granulomas are compatible with tuberculosis, but the non-caseating pattern and lack of TB-specific evidence weakens this diagnosis. \\
Gastric ulcer (K25) & tier 1 / 0.333 & Gastric ulcers are present, but ulcer disease alone does not explain the granulomatous pathology as the final diagnosis.
\end{tabular}
\\
\midrule
\textit{GuideSkill} Prediction & Sarcoidosis (D86), selected after fusion. The tier-3 sarcoidosis evidence overrides the LLM's top-ranked Crohn's disease prediction. \\
\bottomrule
\end{tabular}
\caption{Case study on MedCaseReasoning. The table includes the patient case, ground-truth diagnosis, the LLM's top-5 differential diagnosis, and the executable skill trace used by \textit{GuideSkill}.}
\label{tab:case_study_ddx}
\end{table*}

\begin{figure*}[t]
\begin{promptbox}{Executed Skill: Sarcoidosis (D86)}
def score_sarcoidosis(case):
    """id: 148 | icd10: D86 | disease: Sarcoidosis"""

    # DIRECT (tier 3): biopsy non-caseating granuloma + mimics excluded,
    # OR three-pillar clinical/radiographic/pathologic support,
    # OR cardiac/neuro imaging support with mimics excluded.
    direct_confirmed = (
        (biopsy_noncaseating_granuloma is True and mimics_excluded is True)
        or (three_pillars_met is True)
        or (cardiac_neuro_imaging_supported is True and mimics_excluded is True)
    )

    # CONTRADICTING evidence: caseating granuloma, positive AFB/fungal testing,
    # malignant cells, or another confirmed mimic.
    if direct_confirmed and not any_contradiction:
        tier = 3
    elif any_specific:
        tier = 1 if any_contradiction else 2
    elif any_generic:
        tier = 0 if any_contradiction else 1
    else:
        tier = 0

    return ("Sarcoidosis", "D86", tier / 3.0)
\end{promptbox}
\end{figure*}

\begin{figure*}[t]
\begin{promptbox}{Executed Skill: Crohn's Disease (K50)}
def score_crohns_disease(case):
    """id: 313 | icd10: K50 | disease: Crohn's disease [regional enteritis]"""

    # DIRECT (tier 3) requires Crohn's-specific evidence and mimics_excluded.
    direct_findings = [
        (noncaseating_granulomas_with_giant_cells is True) and mimics_excluded,
        (segmental_skip_transmural_terminal_ileum is True) and mimics_excluded,
        (biopsy_proven_crohns is True) and penetrating_or_perianal and mimics_excluded,
    ]

    # SPECIFIC (tier 2): granulomas, skip/transmural pattern, perianal disease,
    # or other characteristic Crohn's features without full direct confirmation.
    if any(direct_findings):
        tier = 3
    elif any_specific:
        tier = 2
    elif any_generic:
        tier = 1
    else:
        tier = 0

    return ("Crohn's disease [regional enteritis]", "K50", tier / 3.0)
\end{promptbox}
\end{figure*}

\clearpage

\section{Prompt Design}
\label{sec:prompt}
The three stages of \textit{GuideSkill} involve multiple LLM calls, with different prompts designed for preprocessing, skill construction, and inference-time reasoning.

In the \textit{Skill Initialization} stage, four prompts are used: \textit{ICD-10 Label Normalization} maps free-text diagnoses to ICD-10 categories; \textit{Guideline to Recommendation} extracts atomic diagnostic recommendations from clinical guidelines; \textit{Recommendation Merge} consolidates recommendations that map to the same ICD-10 category; and \textit{Recommendation to Executable Skill} compiles each merged recommendation into an executable diagnostic skill.

In the \textit{Skill Evolution} stage, three prompts are used: \textit{Rationale Generation} derives diagnosis-supporting rationales from training cases; \textit{Case-to-Recommendation Distillation} distills recurring diagnostic criteria across cases into disease-level recommendations; and \textit{Case-to-Recommendation Refinement} updates existing recommendations with new batches of case evidence.

In the \textit{Skill Execution} stage, three prompts are used: \textit{Differential Diagnosis Proposal} proposes a ranked ICD-10 differential diagnosis; \textit{Skill Feature Grounding} maps a patient case into the input features required by a selected skill; and \textit{Efficient Skill Feature Grounding} grounds features for multiple candidate skills in a single pass, which is used by the efficient variant of \textit{GuideSkill}.

The full prompt templates are shown below. Placeholders to be replaced at runtime are enclosed in curly braces (\{ \}).

\clearpage

\begin{figure*}[t]
\begin{promptbox}{Skill Initialization: ICD-10 Label Normalization}
# Task
You map a free-text clinical diagnosis to a single standard ICD-10 disease CATEGORY.

# Diagnosis
{answer}

# Rules
- Map the diagnosis to its standard ICD-10 category and respond with ONLY a JSON object:
  {"answer": "<category title>", "icd10": "<CODE>"}.
- "answer" is the official ICD-10 category title, written as a disease name with NO code in it.
- "icd10" MUST be the 3-character ICD-10 CATEGORY code: one letter + two digits, NO decimal.
  Use "K35", never "K35.2".
- If the diagnosis does NOT map to any valid ICD-10 disease category, respond with exactly:
  {"answer": null, "icd10": null}.
- If the diagnosis refers to multiple diseases, symptoms only, procedures, treatments, social history,
  or non-disease concepts, respond with exactly:
  {"answer": null, "icd10": null}.

# Examples
{"answer": "Acute appendicitis", "icd10": "K35"}
{"answer": "Pulmonary tuberculosis", "icd10": "A15"}
\end{promptbox}
\end{figure*}

\begin{figure*}[t]
\begin{promptbox}{Skill Initialization: Guideline to Recommendation}
# Task
Extract atomic diagnostic recommendations from the given clinical practice guideline.

# Clinical Practice Guideline
{guideline}

# Rules
- Extract only recommendations that support disease identification, diagnosis, differential diagnosis,
  or clinically meaningful diagnostic decision-making.
- Each recommendation should describe a finding-to-disease or evidence-to-disease rule.
- Keep concrete clinical criteria, thresholds, tests, signs, symptoms, imaging findings, laboratory values,
  pathology findings, risk factors, and exclusion criteria when they affect diagnosis.
- Do NOT extract general background, epidemiology, treatment-only advice, administrative guidance,
  or recommendations that do not help diagnose a disease.
- Each extracted recommendation must be self-contained and understandable without the full guideline.
- Stay faithful to the source. Do not add, weaken, strengthen, or invent any criterion.

# Output Format
Return ONLY a JSON array. Each item must have:
{
  "disease": "<diagnosed disease name>",
  "icd10": "<3-character ICD-10 category if available, otherwise null>",
  "recommendation": "<self-contained diagnostic recommendation>"
}
\end{promptbox}
\end{figure*}

\begin{figure*}[t]
\begin{promptbox}{Skill Initialization: Recommendation Merge}
# Role
You merge several diagnostic recommendations that all diagnose the SAME disease
(the same ICD-10 category) into ONE coherent, self-contained diagnostic statement.

# Disease
{disease}

# ICD-10 Category
{icd10}

# Recommendations
{recommendations}

# Instructions
- Synthesize ALL diagnostic criteria into one coherent recommendation.
- Combine overlapping criteria, keep every distinct diagnostic criterion, and remove pure repetition.
- Preserve thresholds, test names, signs, symptoms, imaging findings, laboratory findings, pathology,
  exclusion criteria, and differential clues when present.
- Stay faithful: do not add, strengthen, weaken, or invent any criterion or threshold.
- The merged recommendation should be concise but complete enough to support diagnosis.

# Output Format
Return ONLY the merged diagnostic recommendation text.
\end{promptbox}
\end{figure*}

\begin{figure*}[t]
\begin{promptbox}{Skill Initialization: Recommendation to Executable Skill}
# Task
Convert the diagnostic recommendation into an executable Python diagnostic skill.

# Disease
{disease}

# ICD-10 Category
{icd10}

# Diagnostic Recommendation
{recommendation}

# Requirements
Write a Python function named diagnose(case) that takes one dictionary case as input
and returns a dictionary:
{
  "disease": "<disease name>",
  "icd10": "<3-character ICD-10 category>",
  "tier": <0, 1, 2, or 3>,
  "score": <tier divided by 3>,
  "rationale": "<brief explanation>"
}

# Tier Contract
- tier = 0: The case provides no meaningful support, contradicts the disease, or lacks the required evidence.
- tier = 1: The case is compatible with the disease but only has nonspecific or weak supporting evidence.
- tier = 2: The case is strongly suggestive of the disease based on characteristic findings.
- tier = 3: The case confirms the disease using decisive evidence such as definitive imaging,
  pathology, microbiology, diagnostic test, or explicit gold-standard criterion.

# Instructions
- Encode the recommendation as explicit, executable decision logic.
- Use only variables that can be extracted from a patient case.
- Handle missing values safely. Missing evidence should not be treated as positive evidence.
- Preserve the clinical thresholds and diagnostic criteria from the recommendation.
- Include a docstring that documents every expected key in the case dictionary.
- Do not call external APIs, import non-standard libraries, or use hidden state.
- The skill should be comparable across diseases through the same 0--3 tier scale.

# Output Format
Return ONLY valid Python code.
\end{promptbox}
\end{figure*}

\begin{figure*}[t]
\begin{promptbox}{Skill Evolution: Rationale Generation}
# Task
You are given a patient case and its CONFIRMED final diagnosis.
Produce a diagnostic rationale.

# Patient Case
{question}

# Confirmed Final Diagnosis
{disease} (ICD-10 category {icd10})

# Instructions
Produce:
1. key_variables: the clinically IMPORTANT variables from the case. For each variable,
   give its name and VALUE exactly as stated in the case whenever possible.
2. rationale: FIRST state the necessary variables and their values; THEN explain why they
   establish {disease}.

# Output Format
Return ONLY a JSON object:
{
  "key_variables": [
    {"name": "<variable name>", "value": "<value from case>"}
  ],
  "rationale": "<diagnostic rationale>"
}
\end{promptbox}
\end{figure*}

\begin{figure*}[t]
\begin{promptbox}{Skill Evolution: Case-to-Recommendation Distillation}
# Task
Distill recurring diagnostic criteria for the same disease from multiple case rationales.

# Disease
{disease}

# ICD-10 Category
{icd10}

# Case Rationales
{rationales}

# Instructions
- Identify recurring criteria that reliably support the diagnosis across cases.
- Keep clinically meaningful symptoms, signs, labs, imaging findings, pathology findings,
  history, risk factors, and exclusion criteria.
- Drop incidental, case-specific, demographic, or noisy details that do not define the disease.
- Add discriminating features that help distinguish this disease from plausible alternatives
  when they are supported by the rationales.
- Do not invent criteria not supported by the cases.
- Write the result as a reusable diagnostic recommendation for future cases.

# Output Format
Return ONLY the distilled diagnostic recommendation.
\end{promptbox}
\end{figure*}

\begin{figure*}[t]
\begin{promptbox}{Skill Evolution: Case-to-Recommendation Refinement}
# Task
Refine an existing diagnostic recommendation using a new batch of case rationales.

# Disease
{disease}

# ICD-10 Category
{icd10}

# Existing Recommendation
{current_recommendation}

# New Case Rationales
{new_rationales}

# Instructions
- Preserve the useful diagnostic criteria from the existing recommendation.
- Add new recurring criteria only if they are clinically meaningful and supported by the new batch.
- Remove or soften criteria that appear overly specific, incidental, or unsupported.
- Avoid bloating the recommendation with one-off details.
- Keep the recommendation coherent, self-contained, and faithful to the evidence.
- Do not add thresholds, tests, or claims that are not supported by the rationales.

# Output Format
Return ONLY the refined diagnostic recommendation.
\end{promptbox}
\end{figure*}

\begin{figure*}[t]
\begin{promptbox}{Skill Execution: Differential Diagnosis Proposal}
# Task
Give EXACTLY the {k} most likely diagnoses for this patient as ICD-10 3-character categories,
ranked from most to least likely.

# Patient Case
{question}

# Rules
- Each candidate is one ICD-10 3-character category.
- Diagnose at the category level, not the decimal subcode level.
- You MUST return EXACTLY {k} candidates with {k} DIFFERENT categories.
- Prefer specific disease categories that best explain the patient case.
- Do not include symptoms, procedures, treatments, or non-disease concepts as diagnoses.

# Output Format
Reply with ONLY a JSON array of exactly {k} objects:
[
  {"disease": "<ICD-10 category title>", "icd10": "<3-character ICD-10 code>"}
]
\end{promptbox}
\end{figure*}

\begin{figure*}[t]
\begin{promptbox}{Skill Execution: Skill Feature Grounding}
# Task
You are given a patient case and a diagnostic skill function.
Extract the function's input as a JSON object matching the case dict described in the docstring.

# Patient Case
{question}

# Skill Function
{skill_content}

# Instructions
- Read the docstring to identify every key the case dict expects.
- Extract each value from the patient case as faithfully as possible.
- Use null if the case does not state the value.
- Booleans must be true, false, or null.
- Numeric values should be numbers when possible.
- Do not infer unsupported facts.
- Do not add keys that are not requested by the skill docstring.

# Output Format
Return ONLY a JSON object that can be passed directly as the case argument to diagnose(case).
\end{promptbox}
\end{figure*}

\begin{figure*}[t] 
\begin{promptbox}{Skill Execution: Efficient Skill Feature Grounding}
# Task
You are given a patient case and several diagnostic skill functions, one per candidate diagnosis.
Each skill lists its input features and their meaning in the docstring.
In a SINGLE pass, identify which features are clearly PRESENT (true) in the patient case.

# Patient Case
{question}

# Diagnostic Skills
{skills_block}

# Instructions
- For each candidate diagnosis, list ONLY the feature keys that the case shows to be TRUE (present).
- Omit every feature that is false, absent, or not mentioned; any feature you do not list is assumed false or unknown.
- Do not infer unsupported facts.
- Do not add keys that are not requested by the skill docstrings.

# Output Format
Return ONLY a JSON object mapping each candidate's ICD-10 code to the list of its true features,
i.e. {{"<icd10>": ["<true_feature>", ...], ...}} (use [] if none are true),
directly usable to construct the case argument for each diagnose(case).
\end{promptbox}
\end{figure*}







\end{document}